\documentclass[11pt]{article}
\usepackage{amsmath,amssymb,amsthm,bbm,natbib,xcolor,hyperref}
\usepackage{natbib}
\setcitestyle{round}
\usepackage[margin=1in]{geometry}
\usepackage{graphicx}
\usepackage{nicefrac}
\newtheorem{theorem}{Theorem}
\newtheorem{lemma}{Lemma}
\newtheorem{proposition}{Proposition}

\newtheorem{definition}{Definition}
\newtheorem{corollary}{Corollary}

\newcommand{\R}{\mathbb{R}}

\newcommand{\EE}[1]{\mathbb{E}\left[{#1}\right]}

\newcommand{\PP}[1]{\mathbb{P}\left\{{#1}\right\}}

\newcommand{\eqd}{\stackrel{\textnormal{d}}{=}}
\newcommand{\One}[1]{{\mathbbm{1}}\left\{{#1}\right\}}
\newcommand{\one}[1]{{\mathbbm{1}}_{{#1}}}
\newcommand{\iidsim}{\stackrel{\textnormal{iid}}{\sim}}

  \newcommand\independent{\protect\mathpalette{\protect\independenT}{\perp}}
\def\independenT#1#2{\mathrel{\rlap{$#1#2$}\mkern2mu{#1#2}}}

\newcommand{\dtv}{\mathrm{d}_{\mathrm{TV}}}
\newcommand{\alg}{\mathcal{A}}
\newcommand{\Xcal}{\mathcal{X}}
\newcommand{\Ycal}{\mathcal{Y}}
\newcommand{\Zcal}{\mathcal{Z}}

\newcommand{\quantile}{\mathrm{Quantile}}
\newcommand{\Ch}{\widehat{C}_n}
\newcommand{\bZ}{\mathbf{Z}}

\newcommand{\bS}{\mathbf{S}}

\newcommand{\bw}{\mathbf{w}}

\newcommand{\1}{\ensuremath{{\sf (i)}}}
\newcommand{\2}{\ensuremath{{\sf (ii)}}}
\newcommand{\3}{\ensuremath{{\sf (iii)}}}

\begin{document}

\begin{center}

{\LARGE
  \textbf{Predictive inference for time series: 
  why is split conformal effective despite temporal dependence?}
  
  }

\vspace*{.2in}

{\large{
\begin{tabular}{ccc}
Rina Foygel Barber$^\star$ and Ashwin Pananjady$^\dagger$
\end{tabular}
}}
\vspace*{.2in}

\begin{tabular}{c}
$^\star$Department of Statistics, University of Chicago \\
$^\dagger$Schools of Industrial and Systems Engineering and
Electrical and Computer Engineering, \\
Georgia Tech 
\end{tabular}

\vspace*{.2in}

\today

\vspace*{.2in}

\end{center}

\begin{abstract}
We consider the problem of uncertainty quantification for prediction in a time series: if we use past data to forecast the next time point, can we provide valid prediction intervals around our forecasts? To avoid placing distributional assumptions on the data, in recent years the conformal prediction method has been a popular approach for predictive inference, since it provides distribution-free coverage for any iid or exchangeable data distribution. However, in the time series setting, the strong empirical performance of conformal prediction methods is not well understood, since even short-range temporal dependence is a strong violation of the exchangeability assumption. Using predictors with ``memory''---i.e., predictors that utilize past observations, such as autoregressive models---further exacerbates this problem. In this work, we examine the theoretical properties of split conformal prediction in the time series setting, including the case where predictors may have memory. Our results bound the loss of coverage of these methods in terms of a new ``switch coefficient", measuring the extent to which temporal dependence within the time series creates violations of exchangeability. Our characterization of the coverage probability is sharp over the class of stationary, $\beta$-mixing processes. Along the way, we introduce tools that may prove useful in analyzing other predictive inference methods for dependent data. 
\end{abstract}

\section{Introduction}

Quantifying uncertainty in forecasts is important across many fields, including climate and weather prediction~\citep{eyring2024pushing}, power systems~\citep{cochran2015grid}, and supply chain management~\citep{wen2017multi}. At one extreme, traditional approaches can provide strong theoretical guarantees under parametric assumptions~\citep{box2015time}; however, these approaches can yield misleading conclusions when used alongside black-box ML models, which have become state-of-the-art prediction methods in many time series applications~\citep[e.g.][]{hwang2019improving}. At the other extreme, there exist several black-box uncertainty quantification approaches for time series~\citep{salinas2020deepar,borovykh2017conditional}, but these are difficult to equip with theoretical guarantees.

Conformal prediction methods~\citep{vovk2005algorithmic,shafer2008tutorial} occupy a happy medium between these two extremes, and are often preferred for uncertainty quantification in black-box settings because they are easy to ``wrap around” any existing prediction model while also providing theoretical coverage guarantees~\citep{angelopoulos2023conformal}. In addition to accommodating black-box prediction models, these methods make weak assumptions on the data-generating process, requiring only that the data be exchangeable. Time series data, however, clearly violate these exchangeability assumptions, and a significant body of work has aimed to develop variants of conformal prediction methods that are adapted for the time series setting~\citep[e.g.][]{chernozhukov2018exact,xu2023conformal,gibbs2024conformal}. 

In spite of these developments, the vanilla split conformal algorithm~\citep{papadopoulos2002inductive,lei2018distribution}---without any modifications or constraints on its implementation---remains an appealing choice for uncertainty quantification in time series models because of its low computational cost and effective practical performance~\citep{chernozhukov2018exact,xu2023sequential,oliveira2024split}. On the face of it, this may seem quite surprising: due to temporal dependence, time series data is generally far from exchangeable, so how can a framework whose justification relies on exchangeability perform so well? The purpose of this paper is to explain the (often) strong performance of this algorithm in the time series setting.

\subsection{The predictive inference problem}
To be concrete, suppose we have a time series of covariate-response data $\bZ = (Z_1, \ldots, Z_{n + 1})$, with data points $Z_i=(X_i,Y_i) \in \Xcal\times\Ycal=\Zcal$, where $X_i$ is the feature and $Y_i$ is the response. The data point at index $n + 1$ is considered to be the ``test point'', with $X_{n+1}$ observed but $Y_{n+1}$ unobserved, while for $i\in[n]:=\{1,\dots,n\}$ we observe the labeled point $(X_i,Y_i)$. We wish to perform uncertainty quantification on the test response $Y_{n + 1}$, by providing a prediction interval around some estimated value. For instance, given a pretrained predictive model $\widehat{f}$ (where $\widehat{f}(X_{n+1})$ is our point prediction for $Y_{n+1}$), how can we use the available data $(X_i,Y_i)_{i\in[n]}$ to construct a prediction interval around $\widehat{f}(X_{n+1})$ that is likely to contain the target, $Y_{n+1}$---and, how can we do so without placing overly strong assumptions on the distribution of the data?

Split conformal prediction \citep{papadopoulos2002inductive,vovk2005algorithmic} addresses this problem with the following method. Suppose we have a score function $s: \Zcal \to \mathbb{R}$ that we can evaluate on our data points. Assume for the moment that $s$ is pretrained---that is, the definition of $s$ does not depend on $\bZ$. Treating our first $n$ data points as calibration data, we observe that if all data points are iid, the score evaluated at the test point, $s(Z_{n+1})$, must \emph{conform} to the scores of the calibration data points, $(s(Z_i))_{i\in[n]}$ (in that it must be drawn from the same distribution). If we wish to guarantee coverage with probability at least $1-\alpha$, the split conformal prediction set is then given by
\begin{equation}\label{eqn:C_pretrained}
\Ch(X_{n+1}) = \left\{ y\in\Ycal : s(X_{n+1},y) \leq \quantile_{(1-\alpha)(1+1/n)}(s(Z_1),\dots,s(Z_n))\right\},
\end{equation}
where the correction factor $1 + \nicefrac{1}{n}$ to the coverage is to account for the fact that we can only compute the quantile on the $n$ training points without including the test point.\footnote{To formally define the notation $\quantile(\cdot)$, which computes the quantile of a finite list of values,  for any $v\in\R^m$ we use $\quantile_{b}(v)$ to denote the $\lceil bm \rceil$-th order statistic of the vector, i.e., $v_{(\lceil bm \rceil)}$ where $v_{(1)}\leq \dots \leq v_{(m)}$. We will use the convention that $\quantile_b(v)=\infty$ if $b>1$, and $\quantile_b(v)=-\infty$ if $b\leq 0$.}
A canonical example in the setting of a real-valued response ($\Ycal=\R$) is the regression score, $s(z) = |y-\widehat{f}(x)|$ where $z=(x,y)$ and $\widehat{f}$ is a pretrained regression model. This leads to a prediction set of the form $\Ch(X_{n+1}) = \widehat{f}(X_{n+1})\pm \quantile_{(1-\alpha)(1+1/n)}(s(Z_1),\dots,s(Z_n))$. However, the split conformal method may be implemented with any score function. 

In practice, however, the score function is generally not independent of all observed data. For instance, in the setting of the residual score, the regression model $\widehat{f}$ must itself be estimated, which requires data. In such cases, split conformal prediction is based on training a score function $s$ on a portion of the first $n$ data points, and calibrating it on the remaining portion.
In particular, letting $\mathcal{A}$ denote the (black-box) algorithm used to train the score on the first $n_0$ data points, the prediction set is given by
\begin{equation}\label{eqn:C_split}
\begin{split}
\Ch(X_{n+1}) = \left\{ y\in\Ycal : s(X_{n+1},y) \leq \quantile_{(1-\alpha)(1+1/n_1)}(s(Z_{n_0+1}),\dots,s(Z_n))\right\}\\\textnormal{ where }s = \alg(Z_1,\dots,Z_{n_0})\textnormal{ and }n_1 = n-n_0.\end{split}
\end{equation}
In the setting of exchangeable data, split conformal prediction (with any score function, either pretrained as in~\eqref{eqn:C_pretrained} or data-dependent as in~\eqref{eqn:C_split}) is guaranteed to cover $Y_{n+1}$ with probability at least $1-\alpha$ \citep{papadopoulos2002inductive,vovk2005algorithmic}.

Throughout the paper, we will use the term ``pretrained'' to describe the setting where the function $s$ is independent of the data $\bZ$ (for instance, $s$ uses a model that was trained on an entirely separate dataset), to distinguish it from the scenario where $s$ is trained on $Z_1,\dots,Z_{n_0}$, as in~\eqref{eqn:C_split}. In the setting of iid data, there is essentially no distinction between the pretrained construction~\eqref{eqn:C_pretrained} and the split conformal construction~\eqref{eqn:C_split} (aside from having $n$ versus $n_1$ many calibration points), since either way, the score function $s$ is independent of the calibration data. In contrast, for a time series setting, this is no longer the case: the first few calibration points, $Z_{n_0+i}$ for small $i$, may have high dependence with the score function $s$, since $s$ itself is dependent on all data up to time $n_0$. For this reason, the split conformal setting will require a more careful analysis.

\subsection{A motivating numerical experiment}
Despite the theoretical requirement of exchangeability, split conformal prediction has often been observed to perform well on time series data, where the standard exchangeability assumptions are strongly violated. This phenomenon has been observed repeatedly in the literature, where split conformal is often observed to be competitive with other uncertainty quantification methods for time series~\citep[see, e.g.][]{oliveira2024split,xu2023conformal}.

For concreteness, let us illustrate the coverage of split conformal on a toy example. Let $(W_j)_{j \in \mathbb{Z}}$ denote a collection of standard Gaussian variables, and for each $i \in [n+1]$, set  $\epsilon_i = \sum_{j = i - t}^i W_j$ to be a moving average process of order $t$ with unit coefficients; denote the joint distribution of $(\epsilon_i)_{i\in[n+1]}$ by $\mathsf{MA}(t; \mathbf{1})$. Suppose we have a time series of data $(X_i, Y_i)_{i\in[n+1]}$ generated from the standard regression model
\begin{align} \label{eq:MAR}
Y_i = f(X_i) + \epsilon_i, \quad \text{ where } (\epsilon_i)_{i\in[n+1]} \sim \mathsf{MA}(t; \mathbf{1}).
\end{align}
Now suppose that as a pretrained (and memoryless) predictor, we are given access to the true function $f$, and we use the absolute residual as the score function, i.e. $s(X, Y) = | Y - f(X) |$. With the goal of achieving coverage with probability at least $1 - \alpha$, we then output the pretrained prediction set~\eqref{eqn:C_pretrained}; note that with our choice of score function, this set is an interval. 

\begin{figure}[ht!]
\includegraphics[width=\linewidth]{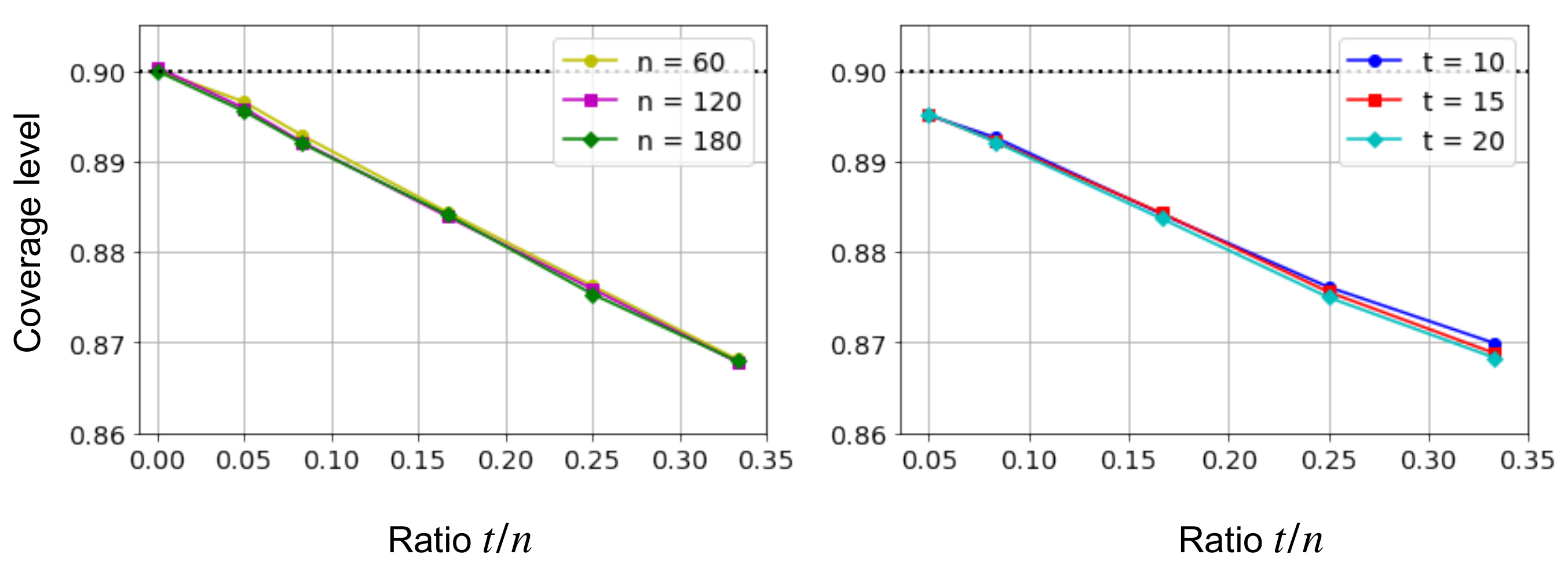}
\caption{Coverage of the pretrained conformal prediction set~\eqref{eqn:C_pretrained} on a sequence of length $n$ from the moving average process~\eqref{eq:MAR} of order $t$. The desired (i.e. nominal) coverage is $90\%$ in both experiments, and is denoted by a dotted line. Each point is generated by averaging over $10^6$ empirical trials.} \label{fig:MA}
\end{figure}

In Figure~\ref{fig:MA}, we plot the coverage achieved by this prediction interval. Clearly, the prediction interval achieves the desired coverage if the MA process has order $t = 0$, in which case the process is iid, but for all other settings it suffers from a modest loss of coverage. Based on these plots, we might conjecture that the loss in coverage for split conformal prediction is proportional to $\nicefrac{t}{n}$. But can we guarantee that the coverage loss is always bounded in this fashion? This paper will provide an affirmative answer to this question for a larger class of time series models, accommodating not just pretrained scores and memoryless predictors but also the split conformal approach~\eqref{eqn:C_split} and predictors with memory, which we introduce next.

\subsection{Pretrained and split conformal for predictors with memory}

Note that it is typical in time series models to use a prediction for response $Y_i$ that does not only depend on the covariate $X_i$ at time $i$, but also on the most recently observed $L$ points. Indeed, equipping a predictor with memory is likely to be effective (i.e., to yield more accurate predictions) precisely when there are dependencies in the time series. In such cases, however, the score function can no longer be thought of as a map from $\Zcal \to \mathbb{R}$, since it is computed using a memory-$L$ predictor. Instead, abusing notation slightly, the score function is now given by a higher dimensional map, $s: \Zcal^{L + 1} \to \mathbb{R}$---for instance, if we have a predictive model $\widehat{f}(x;z_{-1},\dots,z_{-L})$ that predicts the response $y$ given the current feature $x$ in addition to the data from the preceding $L$ time points, we might choose a residual score, $s(z;z_{-1},\dots,z_{-L}) = |y - \widehat{f}(x;z_{-1},\dots,z_{-L})|$, where $z=(x,y)$. The pretrained conformal prediction set is then given by 
\begin{subequations} \label{eqn:C_pretrained_lag}
\begin{equation}
\begin{split}\Ch(&X_{n+1};Z_n,\dots,Z_{n-L+1}) ={}\\ &\left\{ y\in\Ycal : s((X_{n+1},y);Z_n,\dots,Z_{n-L+1}) \leq \quantile_{(1-\alpha)(1+\frac{1}{n-L})}(S_{L+1},\dots,S_n)\right\}\end{split}\end{equation}
where, for each $i=L+1,\dots,n$,
\begin{equation}\label{eqn:define_lag_scores}S_i = s(Z_i;Z_{i-1},\dots,Z_{i-L})\end{equation}
\end{subequations}
is the score for prediction at time $i$ using the previous $L$ time points. In the case $L=0$, this simply reduces back to the original construction~\eqref{eqn:C_pretrained}.
On the other hand, for $L\geq 1$, note that our calibration set only yields $n-L$ many scores $S_{L+1},\dots,S_n$, rather than $n$ scores as before---this is because we cannot evaluate the conformity score for any data point at time $i\leq L$, since we do not have $L$ preceding time points available to make a prediction.

Analogously, the split conformal prediction set is given by
\begin{equation}\label{eqn:C_split_lag}
\begin{split}&\Ch(X_{n+1};Z_n,\dots,Z_{n-L+1}) ={}\\ &\left\{ y\in\Ycal : s((X_{n+1},y);Z_n,\dots,Z_{n-L+1}) \leq \quantile_{(1-\alpha)(1+\frac{1}{n_1-L})}(S_{n_0+L+1},\dots,S_n)\right\},\end{split}\end{equation}
where $n_1=n-n_0$ and the trained score function is given by $s = \alg(Z_1,\dots,Z_{n_0})$ and where $S_i$ is defined as in~\eqref{eqn:define_lag_scores} for each $i=n_0+L+1,\dots,n$. Here again, we have abused notation in defining $\mathcal{A}$ to be a training algorithm that outputs a score function having memory $L$.

\subsection{Related work}

The conformal prediction literature is vast; we refer to the books~\citep{vovk2005algorithmic,angelopoulos2023conformal,angelopoulos2024theoretical} for a comprehensive treatment of the broader literature, and focus this section only on theoretically grounded conformal prediction methods for time series.

\paragraph{Existing results explaining conformal prediction on time series.} Since our focus is on explaining why split conformal is effective on time series data, we begin by surveying existing explanations for why conformal prediction methods more generally can be effective beyond exchangeability. Most of these explanations are based on defining explicit deviations from exchangeability~\citep{barber2025unifying}. For example,~\cite{barber2023conformal} defined a measure motivated by settings with distribution shift---however, this measure of deviation from exchangeability can be large for time series, since it relies on the time series $\bZ$ having approximately the same distribution if we swap the last data point with an earlier data point, $(Z_1,\dots,Z_{k-1},Z_{n+1},Z_{k+1},\dots,Z_n,Z_k)$ (which, under strong short-term temporal dependence, might in fact substantially change the joint distribution). Other deviations from exchangeability include assumptions that the scores are strongly mixing~\citep{xu2023conformal}, but theoretical guarantees are only provided under the additional condition that the predictor is consistent. Note that we may not have consistent prediction in black-box settings, but would still like valid coverage. Closely related to our work is the recent paper by~\cite{oliveira2024split}, who also study split conformal prediction in time series. Among other results, they show using concentration inequalities for empirical processes that split conformal prediction incurs a loss of coverage on the order $(\mathsf{t_{mix}}/n)^{1/2}$ for a $\beta$-mixing process with mixing time $\mathsf{t_{mix}}$. While this shows that the coverage loss is asymptotically vanishing in $n$, it does not explain the type of behavior seen in Figure~\ref{fig:MA}, where the loss of coverage appears to decay proportionally to $1/n$, and to increase linearly in the proxy $t$ for the mixing time. In that sense, our results should be viewed as yielding sharper analogues of the results in~\cite{oliveira2024split}.

\paragraph{Modifying conformal methods for the time series setting.} Moving beyond split conformal, other methods have been specifically developed for the time series setting (and more broadly for non-exchangeable settings). Notable examples are conformal prediction algorithms due to~\cite{chernozhukov2018exact,chernozhukov2021distributional}, which rely on approximate block exchangeability of time series data and ensemble methods due to~\cite{xu2023conformal}, which are proven to work when we have a consistent predictor. Other methods are based on weighted versions of conformal prediction~\citep{tibshirani2019conformal,fannjiang2022conformal,prinster2024conformal}, but these approaches involve correcting for a known distribution shift---information that is not typically available for most time series data. A final family of methods is derived from online learning~\citep[e.g.,][]{gibbs2021conformal,gibbs2024conformal}, and views the construction of uncertainty sets as a game between nature and the statistician.

\subsection{Contributions and organization}

Our contributions can be summarized as follows:
\begin{itemize}
\item We introduce the notion of a \emph{switch coefficient} for a dependent stochastic process, which measures the total variation distance when we swap certain subvectors of the time series. We show that the switch coefficients can be bounded for $\beta$-mixing processes---and consequently, processes such as the one in the motivating example~\eqref{eq:MAR} are covered by our theory.
\item We bound the coverage loss of pretrained conformal prediction by a function of the switch coefficient of the score process. For the MA process and its relatives, this result theoretically confirms the empirical observation made in Figure~\ref{fig:MA}, and holds over a more general class of stochastic processes while accommodating predictors with memory. Moreover, we show that our characterization is tight over the class of stationary, $\beta$-mixing sequences.
\item We extend these findings to split conformal prediction, showing that even here, the coverage loss is bounded by a related switch coefficient.
\end{itemize}

The rest of this paper is organized as follows. In Section~\ref{sec:setup}, we introduce the switch coefficient of a stochastic process, and show how this relates to standard notions of mixing. Section~\ref{sec:main_results} presents our main results for both pretrained and split conformal prediction. We conclude the main paper with a discussion in Section~\ref{sec:discussion} and postpone our proofs to Appendix~\ref{app:proofs}.

\section{Quantifying dependence in the time series} \label{sec:setup}

In this section, we examine the distribution of the time series of data points $\bZ = (Z_1,\dots,Z_{n+1})$, and define coefficients that measure the extent to which the data violates the exchangeability assumption due to temporal dependence.

\subsection{The switch coefficients}

To begin, we need to define notation for deleting a block of entries from a vector.

\begin{definition}[The deletion operation]\label{def:deletion}
    Fix any $m\geq k\geq 1$, and any $\tau\in\{0,\dots,m-1\}$. Let $\bw= (w_1,\dots,w_m)$ be a vector of length $m$ (taking values in any space). We define $\Delta^0_{k,\tau}(\bw)$ and $\Delta^1_{k,\tau}(\bw)$, which are each subvectors of $\bw$ obtained by deleting $\tau$ many entries, as follows. If $1\leq k\leq m-1-\tau$, we define
    \[\Delta^0_{k,\tau}(\bw) = (w_1,\dots,w_{m-\tau-k},w_{m-k+1},\dots,w_m),\]
    which is the subvector consisting of the \emph{first} $m-\tau-k$ entries of $\bw$ followed by the \emph{last} $k$ entries of $\bw$, and is obtained by deleting a block of $\tau$ many entries after position $m-\tau-k$. Similarly, define
    \[\Delta^1_{k,\tau}(\bw) = (w_{k+\tau+1},\dots,w_m,w_1,\dots,w_k),\]
    which is the subvector consisting of the \emph{last} $m-\tau-k$ entries of $\bw$ followed by the \emph{first} $k$ entries of $\bw$. If instead $m-\tau\leq k\leq m$, then we define
    \[\Delta^0_{k,\tau}(\bw) = (w_{\tau+1},\dots,w_m)\textnormal{ and }
    \Delta^1_{k,\tau}(\bw) = (w_{k-m+\tau+1},\dots,w_k),\]
    which each consist of $m-\tau$ consecutive entries of $\bw$.
\end{definition}
\noindent See Figures~\ref{fig:illustrate_subvectors_1} and~\ref{fig:illustrate_subvectors_2} for an illustration of these definitions. In particular, for every $k$, we note that $\Delta^0_{k,\tau}(\bw)$ is defined so that the last entry of $\bw$ (i.e., $w_m$) is in the last position, while $\Delta^1_{k,\tau}(\bw)$ is defined so that $w_k$ is in the last position.

\begin{figure}[t]\centering
\parbox{\textwidth}{\small
\begin{align*}
    &
\textnormal{$\bw = {}$\big( \fcolorbox{black}{blue!10!white}{\parbox{0.17in}{\centering $w_1$} , \parbox{0.17in}{\centering $w_2$}} , \parbox{0.17in}{\centering $w_3$} , \parbox{0.17in}{\centering $w_4$} , \parbox{0.17in}{\centering $w_5$} , \parbox{0.17in}{\centering $w_6$} , \parbox{0.17in}{\centering $w_7$} , \fcolorbox{black}{red!10!white}{\parbox{0.17in}{\centering $w_8$} , \parbox{0.17in}{\centering $w_9$} , \parbox{0.17in}{\centering $w_{10}$}} \big)} \ \rightsquigarrow \ \textnormal{$\Delta^0_{3,5}(\bw) = {}$\big( \fcolorbox{black}{blue!10!white}{\parbox{0.17in}{\centering $w_1$} , \parbox{0.17in}{\centering $w_2$}}  , \fcolorbox{black}{red!10!white}{\parbox{0.17in}{\centering $w_8$} , \parbox{0.17in}{\centering $w_9$} , \parbox{0.17in}{\centering $w_{10}$}} \big)}\\
&
\textnormal{$\bw = {}$\big( \fcolorbox{black}{red!10!white}{\parbox{0.17in}{\centering $w_1$} , \parbox{0.17in}{\centering $w_2$} , \parbox{0.17in}{\centering $w_3$}} , \parbox{0.17in}{\centering $w_4$} , \parbox{0.17in}{\centering $w_5$} , \parbox{0.17in}{\centering $w_6$} , \parbox{0.17in}{\centering $w_7$} , \parbox{0.17in}{\centering $w_8$} ,  \fcolorbox{black}{blue!10!white}{\parbox{0.17in}{\centering $w_9$} , \parbox{0.17in}{\centering $w_{10}$}} \big)} \ \rightsquigarrow \ \textnormal{$\Delta^1_{3,5}(\bw) = {}$\big( \fcolorbox{black}{blue!10!white}{\parbox{0.17in}{\centering $w_9$} , \parbox{0.17in}{\centering $w_{10}$}}  , \fcolorbox{black}{red!10!white}{\parbox{0.17in}{\centering $w_1$} , \parbox{0.17in}{\centering $w_2$} , \parbox{0.17in}{\centering $w_3$}} \big)}
\end{align*}}
\caption{Illustration of the definition of the subvectors $\Delta^0_{k,\tau}(\bw)$ (top) and $\Delta^1_{k,\tau}(\bw)$ (bottom), for a vector $\bw$ of length $m=10$, in the case $k=3$, $\tau = 5$.}\label{fig:illustrate_subvectors_1}
\bigskip\bigskip

\parbox{\textwidth}{\small
\begin{align*}
    &
\textnormal{$\bw = {}$\big( \parbox{0.17in}{\centering $w_1$} , \parbox{0.17in}{\centering $w_2$} , \parbox{0.17in}{\centering $w_3$} , \parbox{0.17in}{\centering $w_4$} , \parbox{0.17in}{\centering $w_5$} , \fcolorbox{black}{red!10!white}{\parbox{0.17in}{\centering $w_6$} , \parbox{0.17in}{\centering $w_7$} , \parbox{0.17in}{\centering $w_8$} , \parbox{0.17in}{\centering $w_9$} , \parbox{0.17in}{\centering $w_{10}$}} \big)} \ \rightsquigarrow \ \textnormal{$\Delta^0_{8,5}(\bw) = {}$\big( \fcolorbox{black}{red!10!white}{\parbox{0.17in}{\centering $w_6$} , \parbox{0.17in}{\centering $w_7$} , \parbox{0.17in}{\centering $w_8$} , \parbox{0.17in}{\centering $w_9$} , \parbox{0.17in}{\centering $w_{10}$}} \big)}\\
&
\textnormal{$\bw = {}$\big( \parbox{0.17in}{\centering $w_1$} , \parbox{0.17in}{\centering $w_2$} , \parbox{0.17in}{\centering $w_3$} , \fcolorbox{black}{red!10!white}{\parbox{0.17in}{\centering $w_4$} , \parbox{0.17in}{\centering $w_5$} , \parbox{0.17in}{\centering $w_6$} , \parbox{0.17in}{\centering $w_7$} , \parbox{0.17in}{\centering $w_8$}} , \parbox{0.17in}{\centering $w_9$} , \parbox{0.17in}{\centering $w_{10}$} \big)} \ \rightsquigarrow \ \textnormal{$\Delta^1_{8,5}(\bw) = {}$\big( \fcolorbox{black}{red!10!white}{\parbox{0.17in}{\centering $w_4$} , \parbox{0.17in}{\centering $w_5$} , \parbox{0.17in}{\centering $w_6$} , \parbox{0.17in}{\centering $w_7$} , \parbox{0.17in}{\centering $w_8$}} \big)}
\end{align*}}
\caption{Illustration of the definition of the subvectors $\Delta^0_{k,\tau}(\bw)$ (top) and $\Delta^1_{k,\tau}(\bw)$ (bottom), for a vector $\bw$ of length $m=10$, in the case $k=8$, $\tau = 5$.}
\bigskip

\label{fig:illustrate_subvectors_2}
\end{figure}

In the results developed in this paper, in order to quantify the extent to which a time series $\bZ\in\Zcal^{n+1}$ fails to satisfy the exchangeability assumption, we will be comparing the distributions of the subvectors $\Delta^0_{k,\tau}(\bZ)$ and $\Delta^1_{k,\tau}(\bZ)$. 
Indeed, in the simple case where the data values $Z_i$ are exchangeable, these subvectors have the same distribution. For instance, if $Z_1,\dots,Z_{n+1}\iidsim P$ for some distribution $P$, then both have the same distribution, $P^{n+1-\tau}$. 
In a time series setting, however, the distributions of these subvectors may differ. The following definition establishes the \emph{switch coefficients}, which compares the distributions of these subvectors---and, as we will see later, characterizes the performance guarantees of split conformal prediction in the time series setting.

\begin{definition}[The switch coefficients]\label{def:Delta}
    Let $n\geq 1$, and let $\bZ\in\Zcal^{n+1}$ be a time series. For each $k\in[n+1]$, define
    \[\Psi_{k,\tau}(\bZ)  = \dtv\big(\Delta^0_{k,\tau}(\bZ),\Delta^1_{k,\tau}(\bZ)\big),\]
    where $\dtv$ denotes the total variation distance, and define 
    \[\bar\Psi_\tau(\bZ) = \frac{1}{n+1}\sum_{k=1}^{n+1}\Psi_{k,\tau}(\bZ).\]
\end{definition}
\noindent Note that while $\Delta^0_{k,\tau}(\bZ)$ and $\Delta^1_{k,\tau}(\bZ)$ are random variables (they each consist of entries of the time series $\bZ$), the switch coefficient $\Psi_{k,\tau}(\bZ)$ is instead a fixed quantity---it is a function of the distribution of $\bZ$, rather than the random variable $\bZ$ itself.

In many practical settings, we might hope that the switch coefficient $\bar\Psi_\tau(\bZ)$ will be small as long as $\tau$ is sufficiently large---that is, while dependence might be strong between consecutive time points, it is plausible that dependence could be relatively weak over a time gap of length $\geq\tau$.

\subsection{Connection to mixing coefficients}
While the switch coefficients are different than the usual conditions appearing in the time series literature, it is straightforward to relate them to a standard mixing condition. Specifically, for a time series $\bZ\in\Zcal^{n+1}$, the $\beta$-mixing coefficient with lag $\tau$ is defined as follows~\citep{doukhan1994mixing}: 
\[
\beta(\tau):= \max_{1 \leq k \leq n-\tau} \dtv\big( (Z_1,\dots,Z_k,Z_{k+\tau+1},\dots,Z_{n+1}), (Z_1,\dots,Z_k,Z'_{k+\tau+1},\dots,Z'_{n+1})\big), \]
where $\bZ'=(Z'_1,\dots,Z'_{n+1})\in\Zcal^{n+1}$ denotes an iid\ copy of $\bZ$. In other words, if $\beta(\tau)$ is small, this means that the subvectors $(Z_1,\dots,Z_k)$ and $(Z_{k+\tau+1},\dots,Z_{n+1})$ are approximately independent. Note that the deletion operation (Definition~\ref{def:deletion}), is closely connected to $\beta$-mixing: the $\beta$-mixing condition is defined in terms of measuring dependence between $(Z_1,\dots,Z_k)$ and $(Z_{k+\tau+1},\dots,Z_{n+1})$, which involves deleting $\tau$ consecutive data points in the time series.

Next we relate $\beta$-mixing coefficients to the switch coefficients defined above.
\begin{proposition}\label{prop:beta_mixing}
    Suppose $\bZ\in\Zcal^{n+1}$ is a stationary time series, with $\beta$-mixing coefficient $\beta(\tau)$.
    Then we have the following bound on the switch coefficients of $\bZ$:
    \begin{align*}
    \begin{cases}
    \Psi_{k,\tau}(\bZ)\leq 2\beta(\tau), \quad &\textnormal{ for $1\leq k\leq n-\tau$},\\
\Psi_{k,\tau}(\bZ)=0, &\textnormal{ for $n-\tau< k\leq n+1$}.\end{cases}
\end{align*} 
\end{proposition}
We prove Proposition~\ref{prop:beta_mixing} in Section~\ref{pf:prop1}.
This result guarantees that any time series with small $\beta$-mixing coefficients must also have small switch coefficients. However, the converse is not true: in particular, as mentioned above, any exchangeable distribution on $\bZ$ ensures $\Psi_{k,\tau}(\bZ)=0$ for all $k,\tau$; however, $\beta(\tau)$ may be large for data that is exchangeable but not iid.

\subsection{Switching data points, or switching scores?}
Suppose we are working with a pretrained score function $s$.
Since the prediction set $\Ch$ depends on the data points only through their scores, we may ask whether the time series of scores is approximately exchangeable. How does this question relate to the properties of the data time series~$\bZ$?

First, consider the simple case $L=0$, with memoryless prediction. Write $\bS = (S_1,\dots,S_{n+1})$ where $S_i = s(Z_i)$ for each $i\in[n+1]$. Since each score $S_i$ is computed as a function of the corresponding data point $Z_i$, it follows by the data processing inequality~\citep[see, e.g.,][Chapter 7]{polyanskiy2025information} that
    \[\Psi_{k,\tau}(\bS)=\dtv(\Delta^0_{k,\tau}(\bS),\Delta^1_{k,\tau}(\bS)) \leq \dtv(\Delta^0_{k,\tau}(\bZ),\Delta^1_{k,\tau}(\bZ))= \Psi_{k,\tau}(\bZ).\]
    Consequently
\[\bar{\Psi}_{\tau}(\bS)\leq \bar{\Psi}_{\tau}(\bZ).\]
In other words, the deviation from exchangeability among the scores, as measured by the averaged switch coefficient $\bar{\Psi}_{\tau}(\bS)$, cannot be higher than the deviation from exchangeability within the time series of data points. Note that in general, it is likely that there is much more dependence among the potentially high-dimensional data points $Z_i$ than among their scores, which are one-dimensional and capture only a limited amount of the information contained in the data. Consequently, in practice $\bar{\Psi}_{\tau}(\bS)$ could be significantly smaller than $\bar{\Psi}_{\tau}(\bZ)$.

In contrast, in the general case with memory $L\geq 0$, the situation is somewhat more complicated. For example, even if the data points $Z_i$ are exchangeable, the scores are \emph{not} exchangeable when the memory $L$ is positive, and indeed may have strong temporal dependence. In particular, writing $\bS = (S_{L+1},\dots,S_{n+1})$ where $S_i = s(Z_i;Z_{i-1},\dots,Z_{i-L})$, we may have $\bar{\Psi}_{\tau}(\bZ)=0$ but $\bar{\Psi}_{\tau}(\bS)>0$, unlike in the memoryless case. Nonetheless, we can still relate the switch coefficients of the scores to those of the data, as shown in the following proposition.
\begin{proposition}\label{prop:Z_vs_S}
    Let $s:\Zcal^{L+1}\to\R$ be a pretrained score function with memory $L\geq 0$, and let $\bZ\in\Zcal^{n+1}$ and $\bS\in\R^{n-L+1}$ be defined as above. Then
    \[\Psi_{k,\tau}(\bS) \leq \Psi_{k+L,\tau-L}(\bZ)\]
    for all $k\in[n-L+1]$ and $\tau\in\{L,\dots,n-L\}$, and consequently,
    \[\bar{\Psi}_\tau(\bS)\leq \frac{n+1}{n-L+1}\bar{\Psi}_{\tau-L}(\bZ).\] 
\end{proposition}
\noindent We prove this proposition in Section~\ref{pf:prop2}. Of course, in the memoryless case ($L=0$), it reduces to the above bounds $\Psi_{k,\tau}(\bS)\leq \Psi_{k,\tau}(\bZ)$ and $\bar\Psi_{\tau}(\bS)\leq \bar\Psi_\tau(\bZ)$.

\section{Main results}\label{sec:main_results}
In this section we will present our main results on the coverage properties of conformal prediction in the time series setting. We will begin by analyzing the setting of a pretrained score function $s$, with the main coverage guarantee presented in Section~\ref{sec:coverage_pretrained}, and with some related results explored in Sections~\ref{sec:coverage_pretrained_tight} and~\ref{sec:coverage_pretrained_upperbound}. Then, in Section~\ref{sec:coverage_splitCP}, we will adapt our coverage guarantee to handle the split conformal setting, where the score function $s$ is trained on a portion of the data. In both cases, our results allow for a memory window of any length $L \geq 0$.

\subsection{Coverage guarantee for the pretrained setting}\label{sec:coverage_pretrained}

We begin by considering pretrained conformal prediction, i.e., the prediction set defined in~\eqref{eqn:C_pretrained_lag}. The following theorem shows that this prediction set cannot undercover if the switch coefficients of the scores are small.

\begin{theorem}\label{thm:coverage}
Let $\bZ\in\Zcal^{n+1}$ be a time series of data points, and 
let $s:\Zcal^{L+1}\to\R$ be a pretrained score function with memory $L$, for some $n\geq L\geq 0$.
Then the prediction set $\Ch$ defined in~\eqref{eqn:C_pretrained_lag} satisfies
\[
\PP{Y_{n+1}\in\Ch(X_{n+1};Z_n,\dots,Z_{n-L+1})} \geq 1-\alpha - \min_{\tau\in\{0,\dots,n-L\}}\left\{\frac{\tau}{n-L+1}+\bar{\Psi}_{\tau}(\bS)\right\},\]
where $\bS=(S_{L+1},\dots,S_{n+1})$, for $S_i = s(Z_i;Z_{i-1},\dots,Z_{i-L})$.
\end{theorem}
\noindent Theorem~\ref{thm:coverage}
is proved in Section~\ref{sec:thm1-pf}. 
While Theorem~\ref{thm:coverage} is stated in terms of the switch coefficients of the scores, combining this result with Propositions~\ref{prop:beta_mixing} and~\ref{prop:Z_vs_S} immediately yields the following corollary, which characterizes the coverage in terms of the properties of the time series $\bZ$.
\begin{corollary}\label{cor:coverage}
    In the setting of Theorem~\ref{thm:coverage}, it holds that
    \small
    \[ \PP{Y_{n+1}\in\Ch(X_{n+1};Z_n,\dots,Z_{n-L+1})} \geq 1-\alpha - \min_{\tau\in\{0,\dots,n-2L\}}\left\{\frac{\tau+L}{n-L+1}+\frac{n+1}{n-L+1}\cdot \bar{\Psi}_{\tau}(\bZ)\right\}.\]
    \normalsize
    Moreover, if we also assume that $\bZ$ is stationary and has $\beta$-mixing coefficients $\beta(\tau)$, then
     \begin{align} \label{eq:beta-mixing-guarantee}
 \PP{Y_{n+1}\in\Ch(X_{n+1};Z_n,\dots,Z_{n-L+1})} \geq 1-\alpha - \min_{\tau \in \{0,\dots,n-2L\}} \left\{\frac{\tau+L}{n-L+1} +  2\beta(\tau)\right\}.
 \end{align}
\end{corollary}

At a high level, we can interpret these results as guaranteeing that if the memory satisfies $L\ll n$ and temporal dependence is weak for some $\tau\ll n$, then the prediction set is guaranteed to have coverage at nearly the nominal level $1-\alpha$. We emphasize that this result does not require any modifications to the conformal prediction method; it simply explains why the method 
might perform reasonably well even when substantial temporal dependence is present, as illustrated in Figure~\ref{fig:MA}.

In the special case of iid data, the minimum is achieved for $\tau=0$ since $\beta(0)=0$. We thus recover the  marginal coverage guarantee \mbox{$\PP{Y_{n+1}\in\Ch(X_{n+1})}\geq 1-\alpha - \frac{L}{n-L+1}$}, and in particular for the memoryless case, coverage is at least $1-\alpha$.
In a similar fashion, one can recover the standard conformal guarantee for exchangeable data in the memoryless ($L=0$) setting: setting $\tau = 0$ and noting that $\Psi_{k,\tau}(\bZ) = 0$ for all $k \in [n + 1]$, here again we obtain the familiar guarantee \mbox{$\PP{Y_{n+1}\in\Ch(X_{n+1})}\geq 1-\alpha$}.

To compare with existing results, we begin by noting that standard results for conformal prediction~\citep{shafer2008tutorial,lei2018distribution,angelopoulos2024theoretical} do not allow for memory-based predictors even when the process $\bZ$ is exchangeable, since memory renders the score process $\bS$ non-exchangeable. Thus, there is no analogue of Theorem~\ref{thm:coverage} in the classical literature on pretrained conformal prediction. Among existing results for pretrained conformal prediction in time series settings, the closest to ours are those of~\citet[Theorem 4]{oliveira2024split}, who show that if the pretrained predictor is memoryless, then its coverage loss on a $\beta$-mixing process is bounded on the order\footnote{Note that the result of~\cite{oliveira2024split} is stated with more parameters, but here we have stated a simplified corollary of their result for the pretrained setting, emphasizing dependence on the pair $(\tau, n)$.} $\min_\tau \{ \sqrt{\nicefrac{\tau}{n}} + 2\beta(\tau) \}$, up to logarithmic factors. Comparing with Corollary~\ref{cor:coverage} above, note that we replace the first term with the ``fast rate" $\nicefrac{\tau}{n}$, leading to a stronger guarantee. Concretely, our improvement is obtained by eschewing arguments based on empirical processes and blocking techniques~\citep{yu1994rates,mohri2010stability} and instead introducing a new technique that exploits the stability of the quantile function upon adding and deleting score values.

\subsection{A matching lower bound}\label{sec:coverage_pretrained_tight}

Our main results provide a guarantee that the loss of coverage, as compared to the nominal level $1-\alpha$, can be bounded by the switch coefficients of the scores---and in turn, can therefore be bounded by the $\beta$-mixing coefficients of the time series, as in~\eqref{eq:beta-mixing-guarantee}. A natural question in light of the comparison with prior work given above is whether our bound on the loss of coverage is tight. In the following result, we provide a matching lower bound; for simplicity, we will work in the memoryless setting ($L=0$), and will assume $(1-\alpha)(n+1)$ is an integer.

\begin{theorem} \label{thm:lb}
Fix any $\alpha\in(0,1)$, data space $\Zcal=\Xcal\times\Ycal$, and sample size $n\geq 1$, where $(1-\alpha)(n+1)$ is an integer. For any constant $b\in[0,1]$, there exists a stationary time series $\bZ \in \Zcal^{n+1}$ and a pretrained score function $s: \Zcal\to\R$, for which it holds that
\[\min_{\tau\in\{0,\dots,n\}} \left\{\frac{\tau}{n+1} + 2\beta(\tau)\right\} \leq b,\]
and the prediction set $\Ch$ defined in~\eqref{eqn:C_pretrained} satisfies
\[
\PP{ Y_{n + 1} \in \Ch(X_{n+ 1}) } \leq \left(1-\frac{b}{4}\right)\cdot(1-\alpha) +  \frac{n(n+1)}{2|\Zcal|}.
\]
\end{theorem}
\noindent Theorem~\ref{thm:lb} is proved in Section~\ref{sec:thmlb-pf}.
 In particular, if $|\Zcal|=\infty$ (i.e., at least one of the spaces $\Xcal$ and $\Ycal$ has infinite cardinality), then we obtain the upper bound
\[
\PP{ Y_{n + 1} \in \Ch(X_{n+ 1}) } \leq  (1-\alpha)  - \frac{1-\alpha}{4}\cdot b \leq  (1-\alpha)  - \frac{1-\alpha}{4} \cdot \min_{\tau\in\{0,\dots,n\}} \left\{\frac{\tau}{n+1} + 2\beta(\tau)\right\}.
\]
This implies that the coverage gap in~\eqref{eq:beta-mixing-guarantee} (and hence, the guarantee given in Theorem~\ref{thm:coverage}) is tight up to a factor $\frac{1-\alpha}{4}$. Since it is typical to take $\alpha \leq 1/2$, this factor should be viewed as a universal constant in the range $[\nicefrac{1}{8}, 1]$.

\subsection{Can the conformal prediction set overcover?}\label{sec:coverage_pretrained_upperbound}

Our results above prove that the switch coefficients of $\bS$ (and consequently, the $\beta$-mixing coefficients of $\bZ$) can be used to bound the loss of coverage of the conformal prediction set---and, moreover, these bounds are tight up to a constant, meaning that there exist settings for which the loss of coverage can indeed be this large. But is it possible that, in other settings, the conformal prediction set can \emph{overcover} rather than undercover? That is, in the time series setting, might conformal prediction lead to sets that are too conservative?

We will now see that the switch coefficients can also be used to provide an upper bound on the coverage probability, to guarantee that the conformal prediction set is not overly conservative. 
\begin{theorem}\label{thm:coverage_upperbd}
In the setting of Theorem~\ref{thm:coverage}, assume also that the scores $S_{L+1},\dots,S_{n+1}$ are distinct almost surely. Then
\begin{multline*}
\PP{Y_{n+1}\in\Ch(X_{n+1};Z_n,\dots,Z_{n-L+1})} \\ {}\leq \frac{\lceil(1-\alpha)(n-L+1)\rceil}{n-L+1} +  \min_{\tau\in\{0,\dots,n-L\}}\left\{\frac{\tau}{n-L+1}+\bar{\Psi}_{\tau}(\bS)\right\}.
\end{multline*}
\end{theorem}
\noindent Theorem~\ref{thm:coverage_upperbd} is proved in Section~\ref{pf:thm3}. We note that as a corollary, upper bounds in terms of the properties of the time series $\bZ$ (analogous to Corollary~\ref{cor:coverage}) also follow in this case. While the coverage upper bound in Theorem~\ref{thm:coverage_upperbd} may be conservative for certain settings, it is possible---analogously to Theorem~\ref{thm:lb}---to construct an example that essentially achieves this upper bound.

\subsection{Coverage guarantee for the split conformal setting}\label{sec:coverage_splitCP}
Next, we turn to split conformal prediction, where the score function $s$ is now trained on a portion of the available data, as in~\eqref{eqn:C_split_lag}. Throughout, we will assume that the sample size is split as $n=n_0+n_1$, where $n_0,n_1\geq 1$ and $n_1\geq L$. 
Write $\bS = (S_{n_0+L+1},\dots,S_{n+1})$, the vector of scores on the calibration set together with the test point score, $S_{n+1}=s(Z_{n+1};Z_n,\dots,Z_{n-L+1})$. Define also
\[\bS_{\textnormal{split},\tau_*} = (S_{n_0+L+\tau_*+1},\dots,S_{n+1}),\]
which deletes the first $\tau_*$ scores for some $\tau_*\geq 0$. The motivation for working with this subvector is that, by deleting the first $\tau_*$ scores, we have removed those scores that may have high dependence with $Z_1,\dots,Z_{n_0}$ (and thus, may have high dependence with the trained score function $s$).
Now we state our main result for coverage in this setting.
\begin{theorem}\label{thm:coverage_split}
Consider the split conformal setting, with the first $n_0$ data points used for training the score function and the remaining $n_1=n-n_0$ points used for calibration.
    Then the prediction set $\Ch$ defined in~\eqref{eqn:C_split_lag} satisfies
\small
    \[\PP{Y_{n+1}\in\Ch(X_{n+1};Z_n,\dots,Z_{n-L+1})} \geq 1-\alpha - \min_{\substack{\tau,\tau_*\geq 0\\ \tau+\tau_*\leq n_1-L}}\left\{ \frac{\tau+\alpha\tau_*}{n_1-\tau_*-L+1} + \bar\Psi_{\tau}(\bS_{\textnormal{split},\tau_*})\right\}.\]
    \normalsize
\end{theorem}
\noindent Theorem~\ref{thm:coverage_split} is proved in Section~\ref{pf-coverage_split}. One might ask why we need to work with $\bS_{\textnormal{split},\tau_*}$, rather than $\bS$. Indeed, by choosing $\tau_*=0$, we simply have $\bS_{\textnormal{split},\tau_*}=\bS$, and this result yields
\[\PP{Y_{n+1}\in\Ch(X_{n+1};Z_n,\dots,Z_{n-L+1})} \geq 1-\alpha - \min_{\tau\in\{0,\dots,n_1-L\}}\left\{ \frac{\tau}{n_1-L+1} + \bar\Psi_{\tau}(\bS)\right\},\]
which is identical to the bound established in Theorem~\ref{thm:coverage} for the pretrained setting except with $n_1$ in place of $n$. But, importantly, in the setting of split conformal, this result is no longer meaningful. This is because  $\bar\Psi_{\tau}(\bS)$ may be large in the time series setting, for any choice of $\tau$. For example, taking the memoryless case $L = 0$ for simplicity, for any $k\leq n_1-\tau$ we have
\[\Psi_{k,\tau}(\bS) = \dtv\big( \Delta^0_{k,\tau}(\bS), \Delta^1_{k,\tau}(\bS)\big) \geq \dtv(S_{n_0+1},S_{n_0+k+\tau+1}) = \dtv(s(Z_{n_0+1}),s(Z_{n_0+k+\tau+1})),\]
where the inequality holds since $S_{n_0+1}$ is the first entry of $\Delta^0_{k,\tau}(\bS)$ while $S_{n_0+k+\tau+1}$ is the first entry of $\Delta^1_{k,\tau}(\bS)$. Since the data point $Z_{n_0+1}$ comes immediately after the data $Z_1,\dots,Z_{n_0}$ used for training $s$, it may be the case that $s$ has  higher dependence with $Z_{n_0+1}$ than with a data point $Z_{n_0+k+\tau+1}$ appearing much later in time---and therefore, the total variation distance between these two data points' scores might be large.

To further explore this point, we will now see how this result can be connected to the $\beta$-mixing coefficients of the time series $\bZ$.
This next result is the analogue of Propositions~\ref{prop:beta_mixing} and~\ref{prop:Z_vs_S}, modified for the split conformal setting.
\begin{proposition}\label{prop:switch_coefs_split_CP}
In the setting of Theorem~\ref{thm:coverage_split}, assume also that $\bZ$ is a stationary time series with $\beta$-mixing coefficients $\beta(\tau)$. Then for each $k,\tau,\tau_*$ with $\tau_*\geq 0$, $L\leq \tau\leq n_1-\tau_*$, and $1\leq k\leq n_1-L+1-\tau_*$, it holds that
\[\Psi_{k,\tau}(\bS_{\textnormal{split},\tau_*}) \leq \begin{cases} 2\beta(\tau_*) + 2\beta(\tau-L), & \textnormal{ for $1\leq k\leq n_1-\tau-\tau_*$,}\\ 2\beta(\tau_*), & \textnormal{ for $n_1-\tau-\tau_*< k \leq n_1-L+1-\tau_*$}. \end{cases}\]
\end{proposition}
We prove Proposition~\ref{prop:switch_coefs_split_CP} in Section~\ref{pf:prop3}. As we will see in the proof, the key step is to bound $\Psi_{k,\tau}(\bS_{\textnormal{split},\tau_*})$ using total variation distances of certain subvectors of $\bZ$ (a more complex form of the switch coefficient). Combining this result with Theorem~\ref{thm:coverage_split}, we obtain the following corollary.
\begin{corollary}\label{cor:coverage_split}
    In the setting of Theorem~\ref{thm:coverage_split}, 
    if $\bZ$ is stationary and has $\beta$-mixing coefficients $\beta(\tau)$, then
\small
     \[\PP{Y_{n+1}\in\Ch(X_{n+1};Z_n,\dots,Z_{n-L+1})} \geq 1-\alpha - \!\!\!\!\min_{\substack{\tau,\tau_*\geq 0\\ \tau+\tau_*\leq n_1-2L}} \left\{\frac{\tau+\alpha\tau_* + L}{n_1-\tau_*-L+1} + 2\beta(\tau) + 2\beta(\tau_*)\right\}.
 \]
 \normalsize
\end{corollary}
\noindent For this result to give a meaningful coverage guarantee in the presence of temporal dependence, we see that we need both $\tau$ and $\tau_*$ to be sufficiently large, so that dependence (as captured by the $\beta$-mixing coefficients) is low.

Let us compare again with the result of~\citet[Theorem 4]{oliveira2024split}, who show that if the score function is memoryless, then its coverage loss for split conformal prediction  on a $\beta$-mixing process is bounded (in our notation and up to logarithmic factors) by a term of the order $\min_{\tau,\tau_*} \{\sqrt{\nicefrac{\tau}{n}} + \sqrt{\nicefrac{\tau_*}{n}} + 2\beta(\tau)+ 2\beta(\tau_*)\}$. As before, comparing with Corollary~\ref{cor:coverage_split} above, note that our bound on the coverage loss is tighter, scaling linearly in $\nicefrac{\tau}{n}$ and $\nicefrac{\tau_*}{n}$. Once again, this improvement is a consequence of our new proof technique.

\section{Discussion} \label{sec:discussion}

Motivated by the question of why pretrained and split conformal prediction are effective in spite of temporal dependence, we introduced a new “switch coefficient” to measure the deviation of scores from exchangeability, and showed that the loss of coverage is bounded whenever the score process has small switch coefficient. This covers the class of $\beta$-mixing processes, and improves upon previous characterizations of the coverage loss. We also showed that our characterization of the coverage loss is tight, and can accurately reflect empirically observed behavior in canonical time series models. An important open question is whether these bounds on under- and over-coverage are optimal for \emph{any} predictive method---that is, whether split conformal is optimal for this class of problems, or whether alternative methods (that may use knowledge of the mixing properties of the time series) could improve these bounds.

We believe that our definitions and proof techniques can find broader applications to other conformal prediction methods. In particular, we expect that the switch coefficient of a process can characterize the coverage loss of other methods when applied to time series data. It is also a natural object in its own right, worth studying for general stochastic processes. Our proof technique, which exploits the stability of the quantile function to the addition or deletion of score values, may also lead to a sharp analysis of other conformal prediction methods. It offers an alternative to blocking techniques~\citep{yu1994rates}, which have seen extensive use in analyzing many statistical estimation and inference methods (beyond uncertainty quantification) in other dynamic settings~\citep{mohri2010stability,yang2017statistical,mou2024optimal,pmlr-v291-nakul25a}.

\subsection*{Acknowledgements}
R.F.B. was partially supported by the National Science Foundation via grant DMS-2023109, and by the Office of Naval Research via grant N00014-24-1-2544.
A.P. was supported in part by the National Science Foundation through grants
CCF-2107455 and DMS-2210734, and by research awards from Adobe, Amazon, Mathworks
and Google. The authors thank Hanyang Jiang, Ryan Tibshirani, and Yao Xie for helpful feedback.

\bibliographystyle{abbrvnat}
\bibliography{bib}

\begin{thebibliography}{32}
\providecommand{\natexlab}[1]{#1}
\providecommand{\url}[1]{\texttt{#1}}
\expandafter\ifx\csname urlstyle\endcsname\relax
  \providecommand{\doi}[1]{doi: #1}\else
  \providecommand{\doi}{doi: \begingroup \urlstyle{rm}\Url}\fi

\bibitem[Angelopoulos and Bates(2023)]{angelopoulos2023conformal}
A.~N. Angelopoulos and S.~Bates.
\newblock Conformal prediction: A gentle introduction.
\newblock \emph{Foundations and Trends in Machine Learning}, 16\penalty0 (4):\penalty0 494--591, 2023.

\bibitem[Angelopoulos et~al.(2024)Angelopoulos, Barber, and Bates]{angelopoulos2024theoretical}
A.~N. Angelopoulos, R.~F. Barber, and S.~Bates.
\newblock Theoretical foundations of conformal prediction.
\newblock \emph{arXiv preprint arXiv:2411.11824}, 2024.

\bibitem[Barber and Tibshirani(2025)]{barber2025unifying}
R.~F. Barber and R.~J. Tibshirani.
\newblock Unifying different theories of conformal prediction.
\newblock \emph{arXiv preprint arXiv:2504.02292}, 2025.

\bibitem[Barber et~al.(2023)Barber, Cand{\`e}s, Ramdas, and Tibshirani]{barber2023conformal}
R.~F. Barber, E.~J. Cand{\`e}s, A.~Ramdas, and R.~J. Tibshirani.
\newblock Conformal prediction beyond exchangeability.
\newblock \emph{The Annals of Statistics}, 51\penalty0 (2):\penalty0 816--845, 2023.

\bibitem[Borovykh et~al.(2017)Borovykh, Bohte, and Oosterlee]{borovykh2017conditional}
A.~Borovykh, S.~Bohte, and C.~W. Oosterlee.
\newblock Conditional time series forecasting with convolutional neural networks.
\newblock In \emph{International Conference on Artificial Neural Networks, ICANN 2017}. Springer, 2017.

\bibitem[Box et~al.(2015)Box, Jenkins, Reinsel, and Ljung]{box2015time}
G.~E. Box, G.~M. Jenkins, G.~C. Reinsel, and G.~M. Ljung.
\newblock \emph{Time series analysis: forecasting and control}.
\newblock John Wiley \& Sons, 2015.

\bibitem[Chernozhukov et~al.(2018)Chernozhukov, W{\"u}thrich, and Yinchu]{chernozhukov2018exact}
V.~Chernozhukov, K.~W{\"u}thrich, and Z.~Yinchu.
\newblock Exact and robust conformal inference methods for predictive machine learning with dependent data.
\newblock In \emph{Conference On learning theory}, pages 732--749. PMLR, 2018.

\bibitem[Chernozhukov et~al.(2021)Chernozhukov, W{\"u}thrich, and Zhu]{chernozhukov2021distributional}
V.~Chernozhukov, K.~W{\"u}thrich, and Y.~Zhu.
\newblock Distributional conformal prediction.
\newblock \emph{Proceedings of the National Academy of Sciences}, 118\penalty0 (48):\penalty0 e2107794118, 2021.

\bibitem[Cochran et~al.(2015)Cochran, Denholm, Speer, and Miller]{cochran2015grid}
J.~Cochran, P.~Denholm, B.~Speer, and M.~Miller.
\newblock Grid integration and the carrying capacity of the {US} grid to incorporate variable renewable energy.
\newblock Technical report, National Renewable Energy Lab.(NREL), Golden, CO (United States), 2015.

\bibitem[Doukhan(1994)]{doukhan1994mixing}
P.~Doukhan.
\newblock \emph{Mixing: Properties and examples}, volume~85 of \emph{Lecture Notes in Statistics}.
\newblock Springer-Verlag, New York, 1994.
\newblock ISBN 0-387-94214-9.
\newblock \doi{10.1007/978-1-4612-2642-0}.
\newblock URL \url{https://doi.org/10.1007/978-1-4612-2642-0}.

\bibitem[Eyring et~al.(2024)Eyring, Collins, Gentine, Barnes, Barreiro, Beucler, Bocquet, Bretherton, Christensen, and Dagon]{eyring2024pushing}
V.~Eyring, W.~D. Collins, P.~Gentine, E.~A. Barnes, M.~Barreiro, T.~Beucler, M.~Bocquet, C.~S. Bretherton, H.~M. Christensen, and K.~Dagon.
\newblock Pushing the frontiers in climate modelling and analysis with machine learning.
\newblock \emph{Nature Climate Change}, 14\penalty0 (9):\penalty0 916--928, 2024.

\bibitem[Fannjiang et~al.(2022)Fannjiang, Bates, Angelopoulos, Listgarten, and Jordan]{fannjiang2022conformal}
C.~Fannjiang, S.~Bates, A.~N. Angelopoulos, J.~Listgarten, and M.~I. Jordan.
\newblock Conformal prediction under feedback covariate shift for biomolecular design.
\newblock \emph{Proceedings of the National Academy of Sciences}, 119\penalty0 (43):\penalty0 e2204569119, 2022.

\bibitem[Gibbs and Cand{\`e}s(2021)]{gibbs2021conformal}
I.~Gibbs and E.~Cand{\`e}s.
\newblock Adaptive conformal inference under distribution shift.
\newblock In \emph{Advances in Neural Information Processing Systems}, volume~34, pages 1660--1672, 2021.

\bibitem[Gibbs and Cand{\`e}s(2024)]{gibbs2024conformal}
I.~Gibbs and E.~J. Cand{\`e}s.
\newblock Conformal inference for online prediction with arbitrary distribution shifts.
\newblock \emph{Journal of Machine Learning Research}, 25\penalty0 (162):\penalty0 1--36, 2024.

\bibitem[Hwang et~al.(2019)Hwang, Orenstein, Cohen, Pfeiffer, and Mackey]{hwang2019improving}
J.~Hwang, P.~Orenstein, J.~Cohen, K.~Pfeiffer, and L.~Mackey.
\newblock Improving subseasonal forecasting in the western {US} with machine learning.
\newblock In \emph{Proceedings of the 25th ACM SIGKDD International Conference on Knowledge Discovery \& Data Mining}, pages 2325--2335, 2019.

\bibitem[Lei et~al.(2018)Lei, G’Sell, Rinaldo, Tibshirani, and Wasserman]{lei2018distribution}
J.~Lei, M.~G’Sell, A.~Rinaldo, R.~J. Tibshirani, and L.~Wasserman.
\newblock Distribution-free predictive inference for regression.
\newblock \emph{Journal of the American Statistical Association}, 113\penalty0 (523):\penalty0 1094--1111, 2018.

\bibitem[Mohri and Rostamizadeh(2010)]{mohri2010stability}
M.~Mohri and A.~Rostamizadeh.
\newblock Stability bounds for stationary $\varphi$-mixing and $\beta$-mixing processes.
\newblock \emph{Journal of Machine Learning Research}, 11\penalty0 (2), 2010.

\bibitem[Mou et~al.(2024)Mou, Pananjady, Wainwright, and Bartlett]{mou2024optimal}
W.~Mou, A.~Pananjady, M.~J. Wainwright, and P.~L. Bartlett.
\newblock Optimal and instance-dependent guarantees for {M}arkovian linear stochastic approximation.
\newblock \emph{Mathematical Statistics and Learning}, 7\penalty0 (1):\penalty0 41--153, 2024.

\bibitem[Nakul et~al.(2025)Nakul, Muthukumar, and Pananjady]{pmlr-v291-nakul25a}
M.~Nakul, V.~Muthukumar, and A.~Pananjady.
\newblock Estimating stationary mass, frequency by frequency.
\newblock In \emph{Proceedings of Thirty Eighth Conference on Learning Theory}, volume 291 of \emph{Proceedings of Machine Learning Research}, pages 4359--4359. PMLR, 30 Jun--04 Jul 2025.

\bibitem[Oliveira et~al.(2024)Oliveira, Orenstein, Ramos, and Romano]{oliveira2024split}
R.~I. Oliveira, P.~Orenstein, T.~Ramos, and J.~V. Romano.
\newblock Split conformal prediction and non-exchangeable data.
\newblock \emph{Journal of Machine Learning Research}, 25\penalty0 (225):\penalty0 1--38, 2024.

\bibitem[Papadopoulos et~al.(2002)Papadopoulos, Proedrou, Vovk, and Gammerman]{papadopoulos2002inductive}
H.~Papadopoulos, K.~Proedrou, V.~Vovk, and A.~Gammerman.
\newblock Inductive confidence machines for regression.
\newblock In \emph{European Conference on Machine Learning}, pages 345--356. Springer, 2002.

\bibitem[Polyanskiy and Wu(2025)]{polyanskiy2025information}
Y.~Polyanskiy and Y.~Wu.
\newblock \emph{Information theory: From coding to learning}.
\newblock Cambridge university press, 2025.

\bibitem[Prinster et~al.(2024)Prinster, Stanton, Liu, and Saria]{prinster2024conformal}
D.~Prinster, S.~D. Stanton, A.~Liu, and S.~Saria.
\newblock Conformal validity guarantees exist for any data distribution (and how to find them).
\newblock In \emph{International Conference on Machine Learning}, pages 41086--41118. PMLR, 2024.

\bibitem[Salinas et~al.(2020)Salinas, Flunkert, Gasthaus, and Januschowski]{salinas2020deepar}
D.~Salinas, V.~Flunkert, J.~Gasthaus, and T.~Januschowski.
\newblock Deep{AR}: Probabilistic forecasting with autoregressive recurrent networks.
\newblock \emph{International journal of forecasting}, 36\penalty0 (3):\penalty0 1181--1191, 2020.

\bibitem[Shafer and Vovk(2008)]{shafer2008tutorial}
G.~Shafer and V.~Vovk.
\newblock A tutorial on conformal prediction.
\newblock \emph{Journal of Machine Learning Research}, 9\penalty0 (3), 2008.

\bibitem[Tibshirani et~al.(2019)Tibshirani, Barber, Cand{\`e}s, and Ramdas]{tibshirani2019conformal}
R.~J. Tibshirani, R.~F. Barber, E.~Cand{\`e}s, and A.~Ramdas.
\newblock Conformal prediction under covariate shift.
\newblock \emph{Advances in neural information processing systems}, 32, 2019.

\bibitem[Vovk et~al.(2005)Vovk, Gammerman, and Shafer]{vovk2005algorithmic}
V.~Vovk, A.~Gammerman, and G.~Shafer.
\newblock \emph{Algorithmic learning in a random world}.
\newblock Springer, 2005.

\bibitem[Wen et~al.(2017)Wen, Torkkola, Narayanaswamy, and Madeka]{wen2017multi}
R.~Wen, K.~Torkkola, B.~Narayanaswamy, and D.~Madeka.
\newblock A multi-horizon quantile recurrent forecaster.
\newblock \emph{arXiv preprint arXiv:1711.11053}, 2017.

\bibitem[Xu and Xie(2023{\natexlab{a}})]{xu2023conformal}
C.~Xu and Y.~Xie.
\newblock Conformal prediction for time series.
\newblock \emph{IEEE transactions on pattern analysis and machine intelligence}, 45\penalty0 (10):\penalty0 11575--11587, 2023{\natexlab{a}}.

\bibitem[Xu and Xie(2023{\natexlab{b}})]{xu2023sequential}
C.~Xu and Y.~Xie.
\newblock Sequential predictive conformal inference for time series.
\newblock In \emph{Proceedings of the 40th International Conference on Machine Learning}, ICML'23, 2023{\natexlab{b}}.

\bibitem[Yang et~al.(2017)Yang, Balakrishnan, and Wainwright]{yang2017statistical}
F.~Yang, S.~Balakrishnan, and M.~J. Wainwright.
\newblock Statistical and computational guarantees for the {B}aum--{W}elch algorithm.
\newblock \emph{Journal of Machine Learning Research}, 18\penalty0 (125):\penalty0 1--53, 2017.

\bibitem[Yu(1994)]{yu1994rates}
B.~Yu.
\newblock Rates of convergence for empirical processes of stationary mixing sequences.
\newblock \emph{The Annals of Probability}, pages 94--116, 1994.

\end{thebibliography}

\appendix

\section{Proofs} \label{app:proofs}

We prove our four main theorems in the first four subsections of this appendix. Proofs of propositions and lemmas can be found in the later subsections.

\subsection{Proof of Theorem~\ref{thm:coverage}} \label{sec:thm1-pf}

By definition of the prediction set~\eqref{eqn:C_pretrained_lag}, the coverage event $Y_{n+1}\in\Ch(X_{n+1};Z_n,\dots,Z_{n-L+1})$ holds if and only if
\[S_{n+1} \leq \quantile_{(1-\alpha)(1+\frac{1}{n-L})}(S_{L+1},\dots,S_n).\]
By properties of the quantile of a finite list (see, e.g., \citet[Lemma 3.4]{angelopoulos2024theoretical}), this event can equivalently be written as
\[S_{n+1} \leq \quantile_{1-\alpha}(\bS).\]

Now fix any $\tau\in\{0,\dots,n-L\}$.
Below, we will show that for each $i\in\{L+1,\dots,n+1\}$, it holds that
\begin{equation}\label{eqn:compare_score_probs_lb}
\PP{S_{n+1} \leq \quantile_{1-\alpha}(\bS)} \geq \PP{S_i \leq \quantile_{1-\alpha - \frac{\tau}{n-L+1}}(\bS)} - \Psi_{i-L,\tau}(\bS).
\end{equation}
Assuming for the moment that this is true, we then calculate
\begin{align*}
    &\PP{S_{n+1} \leq \quantile_{1-\alpha}(\bS)}\\
    &\geq \frac{1}{n-L+1}\sum_{i=L+1}^{n+1}\left[\PP{S_i \leq \quantile_{1-\alpha - \frac{\tau}{n-L+1}}(\bS)} - \Psi_{i-L,\tau}(\bS)\right]\\
    &=\EE{\frac{1}{n-L+1}\sum_{i=L+1}^{n+1}\One{S_i \leq \quantile_{1-\alpha - \frac{\tau}{n-L+1}}(\bS)}} - \frac{1}{n-L+1}\sum_{i=L+1}^{n+1}\Psi_{i-L,\tau}(\bS)\\
    &\overset{\1}{\geq} \left(1-\alpha - \frac{\tau}{n-L+1}\right)- \frac{1}{n-L+1}\sum_{i=L+1}^{n+1}\Psi_{i-L,\tau}(\bS)\\
    &= \left(1-\alpha - \frac{\tau}{n-L+1}\right) - \bar\Psi_{\tau}(\bS),
\end{align*}
where step $\1$ holds since, for any vector $\bw=(w_1,\dots,w_m)\in\R^m$ and any $a\in[0,1]$, it must hold that $\frac{1}{m}\sum_{i=1}^m\One{w_i\leq \quantile_{1-a}(\bw)} \geq 1-a$, by definition of the quantile. Therefore, we have proved the desired lower bound on coverage.

It remains to be shown that~\eqref{eqn:compare_score_probs_lb} holds, for all $i$.
For every $k\in[n-L+1]$, since $\Delta^0_{k,\tau}(\bS)$ and $\Delta^1_{k,\tau}(\bS)$ are each subvectors of $\bS \in \mathbb{R}^{n - L + 1}$, obtained by deleting exactly $\tau$ many entries, it holds surely that 
\begin{equation}\label{eqn:convert_quantile}\quantile_{(1-a)\cdot \frac{n-L+1-\tau}{n-L+1}}(\bS)\leq \quantile_{1-a}\big(\Delta^j_{k,\tau}(\bS)\big)\leq \quantile_{1-a\cdot\frac{n-L+1-\tau}{n-L+1}}(\bS),\end{equation}
for each $j=0,1$, by definition of the quantile. (Recall that we interpret $\quantile_t(\bw)$ as $\infty$ if $t>1$.) In other words, the quantile function is stable to insertion and deletion.
Therefore, for any $k$, we may lower bound the probability of coverage as
\begin{align*}
    &\PP{S_{n+1} \leq \quantile_{1-\alpha}(\bS)}\\
    &\overset{\1}{\geq} \PP{S_{n+1} \leq \quantile_{1-\alpha \cdot \frac{n-L+1}{n-L+1-\tau}}\big(\Delta^0_{k,\tau}(\bS)\big)} \\
    &\overset{\2}{\geq} \PP{S_{L+k} \leq \quantile_{1-\alpha \cdot \frac{n-L+1}{n-L+1-\tau}}\big(\Delta^1_{k,\tau}(\bS)\big)} - \dtv\big(\Delta^0_{k,\tau}(\bS),\Delta^1_{k,\tau}(\bS)\big)\\
    &\overset{\3}{\geq} \PP{S_{L+k} \leq \quantile_{1-\alpha - \frac{\tau}{n-L+1}}(\bS)} - \dtv\big(\Delta^0_{k,\tau}(\bS),\Delta^1_{k,\tau}(\bS)\big).
\end{align*}
Here, steps $\1$ and $\3$ apply~\eqref{eqn:convert_quantile}, while for step $\2$, we use the fact that $S_{n+1}$ is the last entry of $\Delta^0_{k,\tau}(\bS)$ while $S_{L+k}$ is the last entry of $\Delta^1_{k,\tau}(\bS)$. Concretely, both expressions are calculating the probability of the same event (that the last entry is no larger than the quantile), for $\Delta^0_{k,\tau}(\bS)$ and for $\Delta^1_{k,\tau}(\bS)$, which are in turn close in total variation. Finally, taking $k = i-L$, we have verified~\eqref{eqn:compare_score_probs_lb} since $\dtv\big(\Delta^0_{k,\tau}(\bS),\Delta^1_{k,\tau}(\bS)\big) = \Psi_{k,\tau}(\bS) = \Psi_{i-L,\tau}(\bS)$.

\subsection{Proof of Theorem~\ref{thm:lb}} \label{sec:thmlb-pf}

Choose a positive integer $K\leq |\Zcal|$, and let $z_0,\dots,z_{K-1}$ be distinct points in $\Xcal\times\Ycal$. 
We first define two distributions:
\begin{itemize}
    \item Let $P_{\textnormal{cyclic}}$ be a distribution on $\Zcal^{n+1}$, defined as follows.
    Sample $J_1\sim\textnormal{Unif}(\{0,\dots,K-1\})$, and let $J_{i+1} = (J_i+1)\mod K$, for each $i=1,\dots,n$, then return the sequence $(z_{J_1},\dots,z_{J_{n+1}})$.
    \item Let $Q$ denote the uniform distribution on $\{z_0,\dots,z_{K-1}\}$.
\end{itemize}
Now we define our distribution on the time series $\bZ\in\Zcal^{n+1}$. We draw from the mixture distribution 
\[\bZ \sim \frac{b}{4} \cdot P_{\textnormal{cyclic}} + \left(1-\frac{b}{4}\right)\cdot Q^{n+1}.\]
In words, we sample $\bZ$ from $P_{\textnormal{cyclic}}$ with probability $b/4$; otherwise, we sample each of the $n+1$ data points independently and uniformly at random from the set $\{z_0,\dots,z_{K-1}\}$.

First, observe that this distribution is stationary by construction. Next, for any $\tau\geq 0$, we bound the $\beta$-mixing coefficient. Fix any $k\in[n-\tau]$, and as usual let $\bZ'$ denote an iid\ copy of $\bZ$. 
Let $P_0,P_1,P_2$ denote the marginal distribution of the subvectors $(Z_1,\dots,Z_k,Z_{k+\tau+1},\dots,Z_{n+1})$, $(Z_1,\dots,Z_k)$, and $(Z_{k+\tau+1},\dots,Z_{n+1})$, respectively, under the joint distribution $\bZ\sim P_{\textnormal{cyclic}}$. Then, we have
\begin{align*}(Z_1,\dots,Z_k,Z_{k+\tau+1},\dots,Z_{n+1}) &\sim \frac{b}{4}\cdot P_0 + \left(1-\frac{b}{4}\right)\cdot Q^{n+1-\tau}, \\
(Z_1,\dots,Z_k) &\sim \frac{b}{4}\cdot P_1 + \left(1-\frac{b}{4}\right) \cdot Q^k, \text{ and } \\
(Z'_{k+\tau+1},\dots,Z'_{n+1}) &\sim \frac{b}{4}\cdot P_2 + \left(1-\frac{b}{4}\right) \cdot Q^{n+1-\tau-k}.
\end{align*}

Therefore,
\begin{multline*}
    (Z_1,\dots,Z_k,Z'_{k+\tau+1},\dots,Z'_{n+1})\sim{}\\
    \left(\frac{b}{4}\cdot P_1 + \left(1-\frac{b}{4}\right) \cdot Q^k\right) \times \left( \frac{b}{4}\cdot P_2 + \left(1-\frac{b}{4}\right) \cdot Q^{n+1-\tau-k}\right)\\
    = \left(1 - \left(1-\frac{b}{4}\right)^2\right)\cdot P_3 + \left(1-\frac{b}{4}\right)^2 \cdot Q^{n+1-\tau},
\end{multline*}
for an appropriately defined distribution $P_3$. Consequently, we have
\[\dtv\big( (Z_1,\dots,Z_k,Z_{k+\tau+1},\dots,Z_{n+1}), (Z_1,\dots,Z_k,Z'_{k+\tau+1},\dots,Z'_{n+1})\big)\leq \left(1 - \left(1-\frac{b}{4}\right)^2\right).\]
Since this is true for all $k\in[n-\tau]$, the mixing coefficient is bounded as $\beta(\tau)\leq (1 - (1-\frac{b}{4})^2) \leq \frac{b}{2}$, for any $\tau\geq 0$. Thus $\min_\tau \{\frac{\tau}{n+1} + 2\beta(\tau)\} \leq b$.

Next, we prove the bound on coverage. We first need to specify the score function: define 
\[s(z) = \sum_{k=0}^K k \cdot \one{z=z_k}.\]
In other words, $s(z_k) = k$ for each $k\in\{0,\dots,K-1\}$. We are now ready to calculate the coverage probability when the prediction set is constructed with this pretrained score function.
\begin{itemize}
    \item With probability $b/4$, we draw $\bZ$ from $P_{\textnormal{cyclic}}$, meaning that $Z_i = z_{J_i}$ for each $i$ (so that $s(Z_i)=J_i$), with the indices $J_i$ defined via the cyclic construction. If $J_1 \leq K-1-n$, then we have $J_{i+1}=J_i+1$ for all $i\in[n]$, i.e., $J_{n+1}$ is the largest among all the $J_i$'s---and therefore, $s(Z_{n+1}) > \max_{i\in[n]}s(Z_i)$, which implies coverage does not hold. Therefore, on this event, the probability of coverage is at most $\frac{n}{K}$ (i.e., the probability that, when we sample $J_1\in\{0,\dots,K-1\}$ uniformly at random, we draw $J_1>K-1-n$).
    \item With probability $1-b/4$, we draw $\bZ$ from $Q^{n+1}$. In this case, by construction, we have $s(Z_1),\dots,s(Z_{n+1})\iidsim \textnormal{Unif}(\{0,\dots,K-1\})$. On the event that all $n+1$ scores are distinct, by exchangeability the coverage probability is exactly $1-\alpha$ (recalling that we have assumed that $(1-\alpha)(n+1)$ is an integer). And, the event that there is at least one repeated value has probability bounded by $\frac{n(n+1)}{2K}$. In total, therefore, the probability of coverage in this case is bounded by $1-\alpha + \frac{n(n+1)}{2K}$.
\end{itemize}
Combining the cases, then,
\[\PP{Y_{n+1}\in\Ch(X_{n+1})}
\leq \frac{b}{4}\cdot \frac{n}{K}
+ \left(1-\frac{b}{4}\right)\cdot \left(1-\alpha+ \frac{n(n+1)}{2K}\right).\]
Since $\frac{n(n+1)}{2K}\geq \frac{n}{K}$, this completes the proof.

\subsection{Proof of Theorem~\ref{thm:coverage_upperbd}} \label{pf:thm3}
The proof follows essentially the same argument as the lower bound on coverage, Theorem~\ref{thm:coverage}. Fix any $\tau\in\{0,\dots,n-L\}$.
For each $i\in\{L+1,\dots,n+1\}$, it holds that
\begin{equation}\label{eqn:compare_score_probs_ub}
\PP{S_{n+1} \leq \quantile_{1-\alpha}(\bS)} \leq \PP{S_i \leq \quantile_{1-\alpha + \frac{\tau}{n-L+1}}(\bS)} + \Psi_{i-L,\tau}(\bS).
\end{equation}
The proof of this bound is essentially identical to the proof of the analogous bound~\eqref{eqn:compare_score_probs_lb} in the proof of Theorem~\ref{thm:coverage}, so we omit the details. With this bound in place, we calculate
\begin{align*}
    &\PP{S_{n+1} \leq \quantile_{1-\alpha}(\bS)}\\
    &\leq \frac{1}{n-L+1}\sum_{i=L+1}^{n+1}\left[\PP{S_i \leq \quantile_{1-\alpha +\frac{\tau}{n-L+1}}(\bS)} + \Psi_{i-L,\tau}(\bS)\right]\\
    &=\EE{\frac{1}{n-L+1}\sum_{i=L+1}^{n+1}\One{S_i \leq \quantile_{1-\alpha + \frac{\tau}{n-L+1}}(\bS)}} + \frac{1}{n-L+1}\sum_{i=L+1}^{n+1}\Psi_{i-L,\tau}(\bS)\\
    &= \EE{\frac{1}{n-L+1}\sum_{i=L+1}^{n+1}\One{S_i \leq \quantile_{1-\alpha + \frac{\tau}{n-L+1}}(\bS)}} + \bar\Psi_{\tau}(\bS).
\end{align*}
For any vector $\bw=(w_1,\dots,w_m)\in\R^m$ and any $a\in[0,1]$, if $w_1,\dots,w_m$ are distinct, it must hold that $\frac{1}{m}\sum_{i=1}^m\One{w_i\leq \quantile_{1-a}(\bw)} \leq \frac{\lceil (1-a)m\rceil}{m}$, by definition of the quantile. Therefore,
since we have assumed that the scores $S_{L+1},\dots,S_{n+1}$ are distinct almost surely,
\begin{multline*}\EE{\frac{1}{n-L+1}\sum_{i=L+1}^{n+1}\One{S_i \leq \quantile_{1-\alpha + \frac{\tau}{n-L+1}}(\bS)}} \leq \frac{\Big\lceil\Big(1-\alpha + \frac{\tau}{n-L+1}\Big)(n-L+1)\Big\rceil}{n-L+1} \\=\frac{\left\lceil(1-\alpha)(n-L+1)\right\rceil}{n-L+1} + \frac{\tau}{n-L+1}, \end{multline*}
which completes the proof.

\subsection{Proof of Theorem~\ref{thm:coverage_split}} \label{pf-coverage_split}
As in the proof of Theorem~\ref{thm:coverage}, the coverage event $Y_{n+1}\in\Ch(X_{n+1};Z_n,\dots,Z_{n-L+1})$ holds if and only if
\[S_{n+1}\leq \quantile_{1-\alpha}(\bS).\]
And, since the vectors $\bS_{\textnormal{split},\tau_*}$ and $\bS$ are the same aside from the deleted scores $S_{n_0+L+1},\dots,S_{n_0+L+\tau_*}$, it holds surely that
\[ \quantile_{1-\alpha}(\bS) \geq \quantile_{1-\alpha'}(\bS_{\textnormal{split},\tau_*}),\]
where $\alpha'=\alpha\cdot\frac{n_1-L+1}{n_1-\tau_*-L+1}$,
by a similar calculation to~\eqref{eqn:convert_quantile} in the proof of Theorem~\ref{thm:coverage}. Therefore,
\[\PP{Y_{n+1}\in\Ch(X_{n+1};Z_n,\dots,Z_{n-L+1})} \geq \PP{S_{n+1}\leq \quantile_{1-\alpha'}(\bS_{\textnormal{split},\tau_*})},\]
and from now on we only need to bound the probability on the right-hand side.
The remaining steps are exactly the same as in the proof of Theorem~\ref{thm:coverage}, so we omit the details and only summarize briefly. By an argument similar to the one before, we have
\begin{multline*}\PP{S_{n+1}\leq \quantile_{1-\alpha'}(\bS_{\textnormal{split},\tau_*})}
\geq{}\\ \PP{S_i\leq \quantile_{1-\alpha' - \frac{\tau}{n_1-\tau_*-L+1}}(\bS_{\textnormal{split},\tau_*})} - \Psi_{i-L-n_0-\tau_*,\tau}(\bS_{\textnormal{split},\tau_*})\end{multline*}
for each $i\in\{n_0+L+\tau_*+1,\dots,n+1\}$, and therefore, taking an average over all such indices $i$,
\begin{multline*}\PP{S_{n+1}\leq \quantile_{1-\alpha'}(\bS_{\textnormal{split},\tau_*})} \geq 1-\alpha' - \frac{\tau}{n_1-\tau_*-L+1} \\{}- \frac{1}{n_1+1-L-\tau_*}\sum_{i=n_0+L+\tau_*+1}^{n+1}\Psi_{i-L-n_0-\tau_*,\tau}(\bS_{\textnormal{split},\tau_*}).\end{multline*}
Substituting for $\alpha'$ in terms of $\alpha$ and simplifying, this yields the desired bound.

\subsection{Proof of Proposition~\ref{prop:beta_mixing}} \label{pf:prop1}
    First, for any $k>n-\tau$, since the time series is stationary it holds that
    \[\Delta^0_{k,\tau}(\bZ) = (Z_{\tau+1},\dots,Z_{n+1})\eqd (Z_{k+\tau-n},\dots,Z_k) = \Delta^1_{k,\tau}(\bZ)\]
    (where $\eqd$ denotes equality in distribution),
    and therefore $\Psi_{k,\tau}(\bZ) = \dtv\big(\Delta^0_{k,\tau}(\bZ),\Delta^1_{k,\tau}(\bZ)\big) =0$. 
    
    Now we consider the case $k\leq n-\tau$.
    Let $\bZ'=(Z'_1,\dots,Z'_{n+1})\in\Zcal^{n+1}$ denote an iid\ copy of $\bZ$, and define
    \[\widetilde{\bZ}^0 =  (Z_1,\dots,Z_{n+1-\tau-k},Z'_{n+2-k},\dots,Z'_{n+1})\]
    and
    \[\widetilde{\bZ}^1 =  (Z_{k+\tau+1},\dots,Z_{n+1},Z'_1,\dots,Z'_k).\]
    By the triangle inequality, we have
    \[\Psi_{k,\tau}(\bZ) = \dtv\big(\Delta^0_{k,\tau}(\bZ),\Delta^1_{k,\tau}(\bZ)\big)\leq  \dtv\big(\Delta^0_{k,\tau}(\bZ),\widetilde{\bZ}^0\big) +  \dtv\big(\Delta^1_{k,\tau}(\bZ),\widetilde{\bZ}^1\big) + \dtv\big(\widetilde{\bZ}^0,\tilde{\bZ}^1\big) .\]
     Note that by stationarity of $\bZ$ and $\bZ'$, together with independence $\bZ\independent\bZ'$, it holds that $\widetilde{\bZ}^0\eqd \widetilde{\bZ}^1$, and so the last term in the bound above is zero---that is,
     \[ \Psi_{k,\tau}(\bZ) \leq  \dtv\big(\Delta^0_{k,\tau}(\bZ),\widetilde{\bZ}^0\big) +  \dtv\big(\Delta^1_{k,\tau}(\bZ),\widetilde{\bZ}^1\big).\]
    But each of these two remaining terms on the right-hand side is bounded by $\beta(\tau)$ by the definition of $\beta$-mixing, which completes the proof.

\subsection{Proof of Proposition~\ref{prop:Z_vs_S}} \label{pf:prop2}
    First consider the case $1\leq k \leq n-L-\tau$, so that we have
    \[\Delta^0_{k,\tau}(\bS) = (S_{L+1},\dots,S_{n+1-k-\tau}, S_{n+2-k},\dots,S_{n+1})\]
and
\[\Delta^1_{k,\tau}(\bS) = (S_{L+k+\tau+1},\dots,S_{n+1},S_{L+1},\dots,S_{L+k}).\]
Define the function $f_k:\Zcal^{n+L+1-\tau}\to\R^{n-L+1-\tau}$ as
\begin{multline*}(z_1,\dots,z_{n+1-k-\tau},z'_1,\dots,z'_{L+k}) \mapsto {}\\\big(s(z_{L+1};z_L,\dots,z_1), \dots, s(z_{n+1-k-\tau};z_{n-k-\tau},\dots,z_{n-k-\tau-L+1}),\\ s(z'_{L+1};z'_L,\dots,z'_1),\dots,s(z'_{L+k};z'_{L+k-1},\dots,z'_k)\big).\end{multline*}
Then, by construction, $\Delta^0_{k,\tau}(\bS) = f_k\big(\Delta^0_{k+L,\tau-L}(\bZ)\big)$
and $\Delta^1_{k,\tau}(\bS) = f_k\big(\Delta^1_{k+L,\tau-L}(\bZ)\big)$. Therefore,
\[\Psi_{k,\tau}(\bS) = \dtv(\Delta^0_{k,\tau}(\bS),\Delta^1_{k,\tau}(\bS))\leq \dtv(\Delta^0_{k+L,\tau-L}(\bZ),\Delta^1_{k+L,\tau-L}(\bZ)) =\Psi_{k+L,\tau-L}(\bZ),\]
where the inequality follows by data processing.

Next, if $n-L-\tau< k\leq n-L+1$, we have
    \[\Delta^0_{k,\tau}(\bS) = (S_{L+\tau+1},\dots,S_{n+1})\]
and
\[\Delta^1_{k,\tau}(\bS) = (S_{k+2L+\tau-n},\dots,S_{k+L}).\]
In this case, define the function $f_k:\Zcal^{n+L+1-\tau}\to\R^{n-L+1-\tau}$ as
\[(z_1,\dots,z_L,z'_1,\dots,z'_{n+1-\tau}) \mapsto \big(s(z'_{L+1};z'_L,\dots,z'_1), \dots, s(z'_{n+1-\tau};z'_{n-\tau},\dots,z'_{n-L+1-\tau})\big).\]
Then we again have $\Delta^0_{k,\tau}(\bS) = f_k\big(\Delta^0_{k+L,\tau-L}(\bZ)\big)$
and $\Delta^1_{k,\tau}(\bS) = f_k\big(\Delta^1_{k+L,\tau-L}(\bZ)\big)$, and so 
\[\Psi_{k,\tau}(\bS) = \dtv(\Delta^0_{k,\tau}(\bS),\Delta^1_{k,\tau}(\bS))\leq \dtv(\Delta^0_{k+L,\tau-L}(\bZ),\Delta^1_{k+L,\tau-L}(\bZ)) =\Psi_{k+L,\tau-L}(\bZ).\]
Once again, the inequality follows by data processing.

\subsection{Proof of Proposition~\ref{prop:switch_coefs_split_CP}} \label{pf:prop3}

    For each $1\leq k\leq n_1-\tau-\tau_*$, define
    \[\Delta^{\textnormal{split},0}_{k,\tau,\tau_*}(\bZ) = (Z_1,\dots,Z_{n_0},Z_{n_0+\tau_*+1},\dots,Z_{n+1-k-\tau}, Z_{n+2-k},\dots,Z_{n+1})\]
and
\[\Delta^{\textnormal{split},1}_{k,\tau,\tau_*}(\bZ) = (Z_1,\dots,Z_{n_0},Z_{n_0+\tau+\tau_*+k+1},\dots,Z_{n+1},Z_{n_0+\tau_*+1},\dots,Z_{n_0+\tau_*+k}),\]
and for $n_1-\tau-\tau_*< k \leq n_1+1-\tau_*$, define
    \[\Delta^{\textnormal{split},0}_{k,\tau,\tau_*}(\bZ) = (Z_1,\dots,Z_{n_0},Z_{n_0+\tau+\tau_*+1},\dots,Z_{n+1})\]
and
\[\Delta^{\textnormal{split},1}_{k,\tau,\tau_*}(\bZ) =  (Z_1,\dots,Z_{n_0},Z_{n_0+k+\tau+2\tau_*-n_1},\dots,Z_{n_0+k+\tau_*}).\]

The result of the proposition is then an immediate consequence of the following two lemmas.
\begin{lemma}\label{lem:switch_coefs_split_CP_1}
    Under the notation defined above, for any $k,\tau,\tau_*$ with $\tau_*\geq 0$, $L\leq \tau\leq n_1-\tau_*$, and $1\leq k\leq n_1-L+1-\tau_*$, we have
    \[\Psi_{k,\tau}(\bS_{\textnormal{split},\tau_*}) \leq \dtv\big(\Delta^{\textnormal{split},0}_{k+L,\tau-L,\tau_*}(\bZ), \Delta^{\textnormal{split},1}_{k+L,\tau-L,\tau_*}(\bZ)\big).\]
\end{lemma}
\begin{lemma}\label{lem:switch_coefs_split_CP_2}
    Under the notation defined above, if we additionally assume that $\bZ$ is a stationary time series with $\beta$-mixing coefficients $\beta(\tau)$, then for any $k,\tau,\tau_*$ with $\tau,\tau_*\geq 0$, $\tau+\tau_*\leq n$, and $1\leq k\leq n_1 +1 -\tau_*$, we have
\[\dtv\big(\Delta^{\textnormal{split},0}_{k,\tau,\tau_*}(\bZ), \Delta^{\textnormal{split},1}_{k,\tau,\tau_*}(\bZ)\big)\leq \begin{cases} 2\beta(\tau_*) + 2\beta(\tau), & \textnormal{ for $1\leq k\leq n_1-\tau-\tau_*$,}\\ 2\beta(\tau_*), & \textnormal{ for $n_1-\tau-\tau_*< k \leq n_1+1-\tau_*$}. \end{cases}.\]
\end{lemma}

\subsubsection{Proof of Lemma~\ref{lem:switch_coefs_split_CP_1}}

    First suppose $1\leq k\leq n_1-L-\tau-\tau_*$. Then
    \[\Delta^0_{k,\tau}(\bS_{\textnormal{split},\tau_*})
    = (S_{n_0+L+\tau_*+1},\dots,S_{n+1-k-\tau},S_{n+2-k},\dots,S_{n+1})\]
    and
    \[\Delta^1_{k,\tau}(\bS_{\textnormal{split},\tau_*})
    = (S_{n_0+L+\tau_*+k+\tau+1},\dots,S_{n+1},S_{n_0+L+\tau_*+1},\dots,S_{n_0+L+\tau_*+k}).\]
Now define a function
$f_k: \Zcal^{n+L+1-\tau-\tau_*} \to \R^{n_1-L+1-\tau-\tau_*}$ as
\begin{multline*}(z_1,\dots,z_{n_0},z'_1,\dots,z'_{n_1+1-\tau-\tau_*-k},z''_1,\dots,z''_{k+L}) \mapsto {}\\
\big(s(z'_{L+1};z'_L,\dots,z'_1), \dots, s(z'_{n_1+1-\tau-\tau_*-k};z'_{n_1-\tau-\tau_*-k},\dots,z'_{n_1-L-\tau-\tau_*-k+1}), \\
s(z''_{L+1};z''_L,\dots,z''_1),\dots, s(z''_{k+L};z''_{k+L-1},\dots,z''_k)\big)\textnormal{ where }s = \alg(z_1,\dots,z_{n_0}).\end{multline*}
Then we can observe that
\[\Delta^j_{k,\tau}(\bS_{\textnormal{split},\tau_*}) = f_k\big(\Delta^{\textnormal{split},j}_{k+L,\tau-L,\tau_*}(\bZ)\big)\]
for each $j=0,1$. Consequently, by the data processing inequality, we have
\[\Psi_{k,\tau}(\bS_{\textnormal{split},\tau_*}) = \dtv\big(\Delta^0_{k,\tau}(\bS_{\textnormal{split},\tau_*}),\Delta^1_{k,\tau}(\bS_{\textnormal{split},\tau_*})\big)
\leq \dtv\big(\Delta^{\textnormal{split},0}_{k+L,\tau-L,\tau_*}(\bZ),\Delta^{\textnormal{split},1}_{k+L,\tau-L,\tau_*}(\bZ)\big).\]
Next suppose  $n_1-L-\tau-\tau_*< k\leq n_1 -L+1-\tau_*$. Then
    \[\Delta^0_{k,\tau}(\bS_{\textnormal{split},\tau_*})
    = (S_{n_0+L+\tau_*+\tau+1},\dots,S_{n+1})\]
    and
    \[\Delta^1_{k,\tau}(\bS_{\textnormal{split},\tau_*})
    = (S_{n_0+2L+2\tau_*+\tau+k-n_1},\dots,S_{n_0+L+\tau_*+k}).\]
For this case, define the function
$f_k: \Zcal^{n+L+1-\tau-\tau_*} \to \R^{n_1-L+1-\tau-\tau_*}$ as
\begin{multline*}(z_1,\dots,z_{n_0},z'_1,\dots,z'_L,z''_1,\dots,z''_{n_1+1-\tau-\tau_*}) \mapsto {}\\
\big(s(z''_{L+1};z''_L,\dots,z''_1), \dots, s(z''_{n_1+1-\tau-\tau_*};z''_{n_1-\tau-\tau_*},\dots,z''_{n_1-\tau-\tau_*-L+1})\big)\\\textnormal{ where }s = \alg(z_1,\dots,z_{n_0}).\end{multline*}
Then we can observe that
\[\Delta^j_{k,\tau}(\bS_{\textnormal{split},\tau_*}) = f_k\big(\Delta^{\textnormal{split},j}_{k+L,\tau-L,\tau_*}(\bZ)\big)\]
for each $j=0,1$, and so again by data processing we have
\[\Psi_{k,\tau}(\bS_{\textnormal{split},\tau_*})
\leq \dtv\big(\Delta^{\textnormal{split},0}_{k+L,\tau-L,\tau_*}(\bZ),\Delta^{\textnormal{split},1}_{k+L,\tau-L,\tau_*}(\bZ)\big).\]

\subsubsection{Proof of Lemma~\ref{lem:switch_coefs_split_CP_2}}
    First consider the case $1\leq k\leq n_1-\tau-\tau_*$.
    Let $\bZ',\bZ''\in\Zcal^{n+1}$ denote iid\ copies of $\bZ$, and define
    \[\widetilde{\bZ}^0 = (Z_1,\dots,Z_{n_0},Z'_{n_0+\tau_*+1},\dots,Z'_{n+1-k-\tau}, Z''_{n+2-k},\dots,Z''_{n+1})\]
and
\[\widetilde{\bZ}^1 = (Z_1,\dots,Z_{n_0},Z'_{n_0+\tau+\tau_*+k+1},\dots,Z'_{n+1},Z''_{n_0+\tau+1},\dots,Z''_{n_0+\tau+k}),\]
Then, by the triangle inequality,
\[ \dtv\big(\Delta^{\textnormal{split},0}_{k,\tau,\tau_*}(\bZ),\Delta^{\textnormal{split},1}_{k,\tau,\tau_*}(\bZ)\big)\leq  \dtv\big(\Delta^{\textnormal{split},0}_{k,\tau,\tau_*}(\bZ),\widetilde{\bZ}^0\big) 
    {}+ \dtv\big(\Delta^{\textnormal{split},1}_{k,\tau,\tau_*}(\bZ),\widetilde{\bZ}^1\big) + \dtv\big(\widetilde{\bZ}^0,\widetilde{\bZ}^1\big) .\]
Since the three time series $\bZ,\bZ',\bZ''$ are mutually independent and are each stationary, it holds that $\widetilde{\bZ}^0\eqd \widetilde{\bZ}^1$, and so the last term above is zero. Therefore,
     \[\dtv\big(\Delta^{\textnormal{split},0}_{k,\tau,\tau_*}(\bZ),\Delta^{\textnormal{split},1}_{k,\tau,\tau_*}(\bZ)\big)\leq  \dtv\big(\Delta^{\textnormal{split},0}_{k,\tau,\tau_*}(\bZ),\widetilde{\bZ}^0\big) +  \dtv\big(\Delta^{\textnormal{split},1}_{k,\tau,\tau_*}(\bZ),\widetilde{\bZ}^1\big).\]
    Next define
    \[\breve{\bZ}^0 =  (Z_1,\dots,Z_{n_0},Z_{n_0+\tau_*+1},\dots,Z_{n+1-k-\tau},Z''_{n+2-k},\dots,Z''_{n+1}).\]
    Since $\bZ''$ is independent of $\bZ$ and $\bZ'$, we have\begin{multline*}
        \dtv\big(\widetilde{\bZ}^0,\breve{\bZ}^0\big)\\
        =\dtv\big( (Z_1,\dots,Z_{n_0},Z'_{n_0+\tau_*+1},\dots,Z'_{n+1-k-\tau}),(Z_1,\dots,Z_{n_0},Z_{n_0+\tau_*+1},\dots,Z_{n+1-k-\tau})\big)\\
        \overset{\1}{\leq} \beta(\tau_*),
    \end{multline*}
    where step $\1$ holds by definition of $\beta$-mixing. Reasoning similarly, we also have
    \[\dtv\big(\Delta^{\textnormal{split},0}_{k,\tau,\tau_*}(\bZ),\breve{\bZ}^0\big)\leq \beta(\tau),\]
    again by definition of the $\beta$-mixing coefficients. Therefore, again applying the triangle inequality yields
    \[\dtv\big(\Delta^{\textnormal{split},0}_{k,\tau,\tau_*}(\bZ),\widetilde{\bZ}^0\big) 
    \leq \dtv\big(\widetilde{\bZ}^0,\breve{\bZ}^0\big) +\dtv\big(\Delta^{\textnormal{split},0}_{k,\tau,\tau_*}(\bZ),\breve{\bZ}^0\big) \leq \beta(\tau_*) + \beta(\tau). \]
    A similar argument yields that $\dtv\big(\bZ_{\textnormal{split}}^1(k,\tau),\widetilde{\bZ}_{\textnormal{split}}^1(k,\tau)\big)\leq \beta(\tau_*) + \beta(\tau)$, by considering
    \[\breve{\bZ}^1 = (Z_1,\dots,Z_{n_0},Z'_{n_0+\tau+\tau_*+k+1},\dots,Z'_{n+1},Z_{n_0+\tau+1},\dots,Z_{n_0+\tau+k})\]
    in place of $\breve{\bZ}^0$.
    Therefore we have shown that
    \[\dtv\big(\Delta^{\textnormal{split},0}_{k,\tau,\tau_*}(\bZ),\Delta^{\textnormal{split},1}_{k,\tau,\tau_*}(\bZ)\big)\leq 2\beta(\tau_*)+2\beta(\tau).\]

    Next we turn to the case that $n_1-\tau-\tau_*< k \leq n_1+1-\tau_*$. Define
        \[\widetilde{\bZ}^0 = (Z_1,\dots,Z_{n_0},Z'_{n_0+\tau+\tau_*+1},\dots,Z'_{n+1})\]
and
\[\widetilde{\bZ}^1=  (Z_1,\dots,Z_{n_0},Z'_{n_0+k+\tau+2\tau_*-n_1},\dots,Z'_{n_0+k+\tau_*}),\]
    where again $\bZ'$ denotes an iid\ copy of $\bZ$. 
Then, as before,
\[\dtv\big(\Delta^{\textnormal{split},0}_{k,\tau,\tau_*}(\bZ),\Delta^{\textnormal{split},1}_{k,\tau,\tau_*}(\bZ)\big)\leq  \dtv\big(\Delta^{\textnormal{split},0}_{k,\tau,\tau_*}(\bZ),\widetilde{\bZ}^0\big) +  \dtv\big(\Delta^{\textnormal{split},1}_{k,\tau,\tau_*}(\bZ),\widetilde{\bZ}^1\big) + \dtv\big(\widetilde{\bZ}^0,\widetilde{\bZ}^1\big).\]
The first two terms on the right-hand side are each bounded by $\beta(\tau_*)$, by definition of the $\beta$-mixing coefficients, 
while the final term is zero since $\widetilde{\bZ}^0\eqd \widetilde{\bZ}^1$ by stationarity of $\bZ'$, together with the fact that $\bZ\independent\bZ'$. Therefore, for this case we have
    \[\dtv\big(\Delta^{\textnormal{split},0}_{k,\tau,\tau_*}(\bZ),\Delta^{\textnormal{split},1}_{k,\tau,\tau_*}(\bZ)\big)\leq 2\beta(\tau_*),\]
which completes the proof.
\end{document}